\documentclass[11pt]{article}

\usepackage[preprint]{acl}

\usepackage{times}
\usepackage{latexsym}

\usepackage[T1]{fontenc}

\usepackage[utf8]{inputenc}

\usepackage{microtype}

\usepackage{inconsolata}

\usepackage{graphicx}
\usepackage{multirow}
\usepackage{amsmath}
\usepackage{booktabs}
\usepackage[ruled,vlined,linesnumbered]{algorithm2e}
\usepackage[table]{xcolor}

\usepackage{amsthm}

\usepackage{etoolbox} 
\usepackage{mathtools}

\usepackage{booktabs} 
\usepackage[table]{xcolor} 
\usepackage{makecell}

\usepackage{hyperref}
\usepackage[capitalize,noabbrev]{cleveref}
\usepackage[most]{tcolorbox}
\tcbuselibrary{skins,breakable}

\AtBeginDocument{%
  \setlength{\abovedisplayskip}{4pt}
  \setlength{\belowdisplayskip}{4pt}
  \setlength{\abovedisplayshortskip}{2pt}
  \setlength{\belowdisplayshortskip}{2pt}
}

\theoremstyle{plain}

\theoremstyle{definition}

\theoremstyle{remark}

\newcommand{\ours}{\texttt{CompassMem}}

\title{Memory Matters More: Event-Centric Memory as a Logic Map for Agent Searching and Reasoning}

\author{Yuyang Hu, Jiongnan Liu, Jiejun Tan, Yutao Zhu, Zhicheng Dou \\
        Gaoling School of Artificial Intelligence, Renmin University of China\\
        \texttt{yuyang.hu@ruc.edu.cn}, \texttt{dou@ruc.edu.cn}}

\begin{document}
\maketitle
\begin{abstract}
Large language models (LLMs) are increasingly deployed as intelligent agents that reason, plan, and interact with their environments. To effectively scale to long-horizon scenarios, a key capability for such agents is a memory mechanism that can retain, organize, and retrieve past experiences to support downstream decision-making. However, most existing approaches organize and store memories in a flat manner and rely on simple similarity-based retrieval techniques. Even when structured memory is introduced, existing methods often struggle to explicitly capture the logical relationships among experiences or memory units. Moreover, memory access is largely detached from the constructed structure and still depends on shallow semantic retrieval, preventing agents from reasoning logically over long-horizon dependencies. In this work, we propose \ours{}, an event-centric memory framework inspired by Event Segmentation Theory. \ours{} organizes memory as an Event Graph by incrementally segmenting experiences into events and linking them through explicit logical relations. This graph serves as a logic map, enabling agents to perform structured and goal-directed navigation over memory beyond superficial retrieval, progressively gathering valuable memories to support long-horizon reasoning.
Experiments on LoCoMo and NarrativeQA demonstrate that \ours{} consistently improves both retrieval and reasoning performance across multiple backbone models.
\end{abstract}

\section{Introduction}
With the rapid development of large language models (LLMs), agents have evolved from simple interfaces into systems capable of complex reasoning and long-term interaction with environments~\citep{agentmemorysurveyzhang, agentsurveywang}. To support such behaviors, agents require memory mechanisms that go beyond simple text generation capabilities~\citep{reasoningbankouyang, agentmemorysurveyzhang}. Ideally, similar to human memory, agent memory should serve not only as a repository of knowledge, but also as a fundamental infrastructure that supports reasoning, planning, and decision-making~\citep{memoryhuawei, zhang2025memevolvemetaevolutionagentmemory}.

\begin{figure*}[t]
  \centering
  \includegraphics[width=\textwidth]{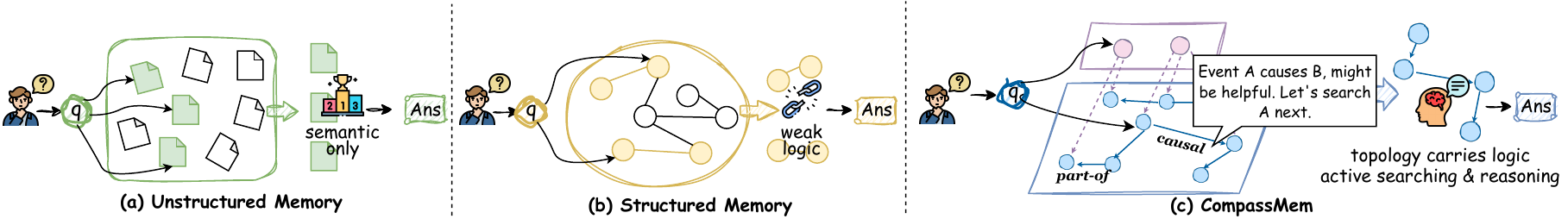}
  \caption{Comparison among \ours{} and the traditional agent memory framework.}
  \label{fig:Intro}
\end{figure*}

Within the broader field of agent memory research, a significant amount of attention has been directed toward \textit{factual memory}~\citep{memgenzhang,hu2025memoryageaiagents}. 
Factual memory refers to an agent’s capacity to manage explicit information about past events, users, and the external environment. Such memory supports context awareness, personalization, and long-horizon tasks.
Despite significant progress in this area, current approaches face two primary limitations. First, regarding memory structure, most methods rely on flat representations where information is stored as independent text segments, as shown in Figure~\ref{fig:Intro} (a)~\citep{hu2025memoryageaiagents}. While some recent studies have explored structured organizations~\citep{amemxu, zeprasmussen, memtreealireza,sun2025hierarchicalmemoryhighefficiencylongterm,camli}, they often fail to capture essential logical relations, such as causality and temporal sequences (Figure~\ref{fig:Intro} (b))~\citep{yang-etal-2025-eventrag}. Second, regarding memory utilization, prior work primarily depends on simple semantic matching~\citep{zeprasmussen}. This reliance limits memory to functioning as a static storage system rather than an active component that guides the reasoning process.

In contrast, human memory is organized hierarchically and connected through rich logical associations rather than as a collection of isolated facts. Cognitive science offers theoretical support for this organization, particularly through Event Segmentation Theory~\citep{baldassano2017discovering,zacks2007event}. According to this theory, humans naturally perceive continuous experience as a series of discrete and meaningful events. These events form the backbone of long-term memory and are encoded with rich temporal and semantic information~\citep{baldassano2017discovering,ezzyat2011constitutes}. This structured organization facilitates efficient retrieval. It enables the brain to selectively access relevant events by navigating a structured network, which helps guide reasoning and planning in new situations~\citep{anderson1983spreading}. Unfortunately, these capabilities are largely absent in existing agent memory systems. This disparity leads to a critical research question: \emph{Can we structure agent memory in a way that mimics human cognitive organization to support search and reasoning beyond isolated facts?}

Inspired by these cognitive principles, we propose \ours{}, an event-centric memory framework that explicitly models logical relations among memory units and leverages this structure to guide agent searching and reasoning. Unlike traditional approaches that store isolated text snippets, \ours{} incrementally constructs an \emph{Event Graph} from experiences through \textbf{event segmentation}, \textbf{relation extraction}, and \textbf{topic evolution}. In this graph, nodes correspond to coherent event units, while edges encode logical dependencies such as causality and temporal order. 
During the inference phase, agents utilize the Event Graph as a \textbf{structured logic map} rather than a flat list. This structure provides directional cues that guide agent searching and reasoning. It allows agents to prioritize relevant information, follow meaningful logical connections, and avoid redundant retrieval. In this manner, memory goes beyond merely supplying content and actively guides the reasoning process to handle complex queries effectively. \ours{} achieves consistent and substantial improvements over strong baselines on LoCoMo and NarrativeQA, particularly on tasks requiring multi-hop and temporal reasoning. These results demonstrate that explicitly encoding logical structure into memory not only improves retrieval quality, but also enables memory to actively support reasoning, rather than serving as a passive knowledge store.

Our contributions are as follows:

(1) We propose \ours{}, an event-centric memory framework that organizes experiences into event units connected by explicit logical relations.

(2) In \ours{}, we design a graph-based memory retrieval mechanism, enabling agents to actively navigate the Event Graph for logic-aware evidence collection, rather than just relying on flat, similarity-based memory access.

(3) We evaluate \ours{} on dialogue and long-document  benchmarks, observing consistent improvements and validating its effectiveness and generality.

\vspace{-4pt}
\section{Related Work}
Memory has been widely regarded as a core capability of intelligent agents~\citep{agentmemorysurveyzhang}. Early systems such as MemGPT~\citep{memgptcharles} manage long-term memory through paging and segmentation mechanisms, which inspired subsequent frameworks including MemOS~\citep{memosli} and MemoryOS~\citep{memoryoskang}. Methods such as Mem0~\citep{mem0prateek} and MemoryBank~\citep{memorybankzhong} follow a RAG-style paradigm, placing greater emphasis on memory organization and lifecycle management.

Also, a growing line of work explores structured memory representations. Representative examples include tree-based designs such as MemTree~\citep{memtreealireza}, graph-based memories like A-Mem~\citep{amemxu}, and more general hierarchical or compositional memory systems~\citep{zeprasmussen,gmemoryzhang,sgmemwu,camli}. These approaches demonstrate the benefit of introducing structure into memory, particularly for improving organization. In parallel, other studies investigate automatic memory management and adaptation from different perspectives~\citep{memoryr1yan,memalphawang}.

While these methods enrich memory management, memory is still largely treated as a passive storage. Our work designs an event-centric memory that explicitly encodes logical structure and actively guides searching and reasoning.

\section{Preliminary}
In this section, we formalize the task setting and introduce the core concepts used in our approach.

We consider an agent operating over a stream of textual observations at time step $t$, denoted as $\mathcal{X}_t = (x_{t,1}, \dots, x_{t,n})$, where each $x_{t,i}$ is a text unit such as a dialogue turn or a narrative sentence. Given the incoming observations and the previously stored memory $\mathcal{M}^{(t-1)}$, the agent updates its memory through a construction process $\Phi$:
\begin{equation}
\mathcal{M}^{(t)} = \Phi(\mathcal{X}_t, \mathcal{M}^{(t-1)}).
\end{equation}
Here, $\Phi$ first extracts a sub-memory $\mathcal{M}_t$ from the current input stream, and then integrates it with the existing memory $\mathcal{M}^{(t-1)}$, yielding the updated memory $\mathcal{M}^{(t)}$.

At inference time, given a query $q \in \mathcal{Q}$, the agent performs query-dependent memory search through a retrieval process $\Psi$:
\begin{equation}
\mathcal{M}^{(t)}\!\mid_q = \Psi(q, \mathcal{M}^{(t)}),
\end{equation}
where $\mathcal{M}^{(t)}\!\mid_q \subseteq \mathcal{M}^{(t)}$ denotes the subset of memory selected for answering the query.

The final response is generated by a conditional generation function:
\begin{equation}
y = \mathcal{F}(q, \mathcal{M}^{(t)}\!\mid_q),
\label{eq:answer}
\end{equation}
which produces an output $y \in \mathcal{Y}$ conditioned on the query and the retrieved memory. Our goal is to design more effective memory construction processes $\Phi$ and retrieval strategies $\Psi$ to support higher-quality generation.

\vspace{-2pt}
\section{Method}
\label{sec:method}
\vspace{-2pt}

\subsection{Overview}
\label{ssec:overview}
\begin{figure*}[t]
  \centering
  \includegraphics[width=\textwidth]{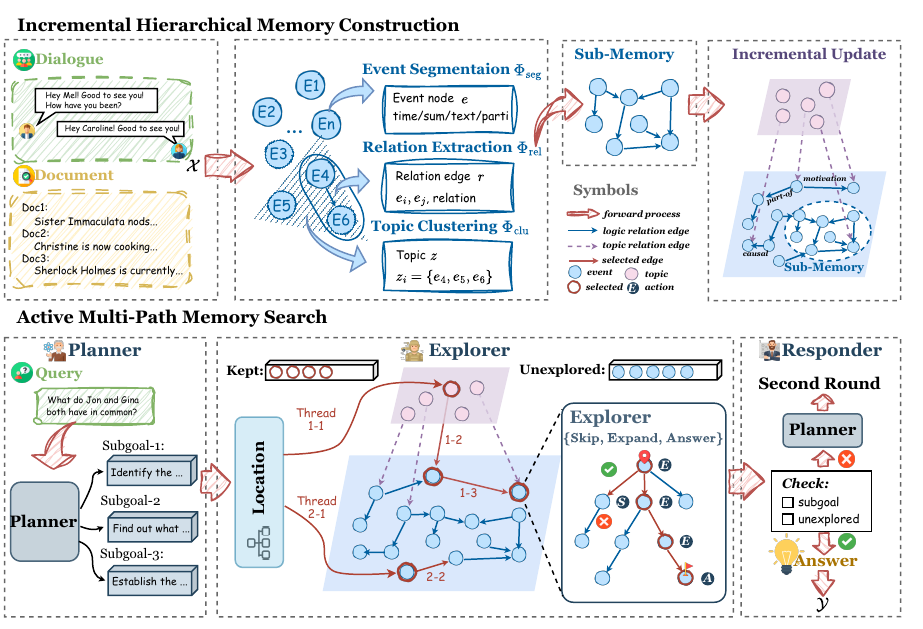}
  \caption{Overview of the proposed \ours{} framework, which contains mainly two part: Incremental Hierarchical Memory Construction and Active Multi-Path Memory Search}

  \label{fig:CompassMem}
\end{figure*}

As shown in~\Cref{fig:CompassMem}, \ours{} is an event-centric memory framework designed to make memory an active guide for agent searching and reasoning. The core idea is to organize  memory as a structured hierarchical Event Graph, where experiences are stored as coherent event units connected by explicit logical relations. 

Memory is constructed incrementally from input streams by segmenting continuous observations into events, extracting relations among them, and integrating the resulting structures into the existing memory over time. During inference, the agent performs logic-aware memory search by actively navigating the Event Graph. Rather than retrieving isolated memories by similarity, the agent follows meaningful logical paths between related events and progressively collects relevant evidence, with the memory structure guiding both where to search and how to reason for complex, long-horizon queries.

\subsection{Incremental Hierarchical Memory Construction}
\label{ssec:method_build}
We construct memory in an incremental manner. The system first segments the input into coherent events, then extracts explicit relations among these events, and finally integrates them into the memory through incremental graph updates.

\noindent\textbf{Event Segmentation}\quad
Event Segmentation Theory (EST)~\citep{baldassano2017discovering} suggests that humans organize continuous experience into discrete and coherent events, which serve as fundamental units of long-term memory. An \emph{event} is not an arbitrary text span, but a meaningful unit obtained by segmenting a continuous experience stream. Following this perspective, we prompt an LLM to identify events from the input stream and extract their attributes:
\begin{equation}
\mathcal{E}_t = \{e_{t_i}\}_{i=1}^{m} = \Phi_{\mathrm{seg}}(\mathcal{X}_t),
\end{equation}
where each event $e_{t_i} \in \mathcal{E}_t$ is represented as
$e_{t_i} = \langle o_{t_i}, \tau_{t_i}, s_{t_i}, \pi_{t_i} \rangle$.
Here, $o_{t_i}$ denotes the span of observations belonging to the event,
$\tau_{t_i}$ captures temporal information,
$s_{t_i}$ is a semantic summary,
and $\pi_{t_i}$ denotes the set of involved participants.

\noindent\textbf{Relation Extraction}\quad
A memory composed of isolated events provides limited support for reasoning~\citep{hu2025memoryageaiagents}. In contrast, humans reason and form associations by following logical connections~\citep{anderson1983spreading}. To enable structured retrieval and multi-step reasoning, we explicitly extract logical relations among event nodes using an LLM-based process:
\begin{align}
\mathcal{R}_t = \{ (e_{t_i}, e_{t_j}, \rho_{t_{ij}}) \} = \Phi_{\mathrm{rel}}(\mathcal{X}_t, \mathcal{E}_t),
\end{align}
where each relation $r_{ij} = (e_i, e_j, \rho_{ij})$ represents a logical dependency between two events. The relation label $\rho_{ij}$ is drawn from an open-ended predicate set $\mathcal{P}$, covering relations such as \emph{causal}, \emph{temporal}, \emph{motivation}, and \emph{part-of}, and allowing new relation types to be introduced as needed. Together, the extracted events $\mathcal{E}_t$ and relations $\mathcal{R}_t$ form the current sub-memory $\mathcal{M}_t = (\mathcal{E}_t,\mathcal{R}_t)$.

\noindent\textbf{Incremental Graph Update}\quad
As memory grows over time, new events must be integrated while preserving coherence, so that newly acquired information can be connected to existing knowledge without introducing redundancy or semantic drift. We incrementally update the memory $\mathcal{M}^{(t-1)}$ by incorporating the new sub-memory $\mathcal{M}_{t}$ through three operations.

\noindent\textbf{Node Fusion \& Expansion}\quad Each new event $e_{\text{new}} \in \mathcal{E}_{t+1}$ is compared against existing events, where $e^{*}$ denotes the most similar existing event.
The integration follows three cases.
If $e_{\text{new}}$ is equivalent to $e^{*}$, the two events are merged.
If a logical relation between $e_{\text{new}}$ and $e^{*}$ is identified, an edge is added to link them.
Otherwise, $e_{\text{new}}$ is inserted as a new node.
This process integrates new information while avoiding redundancy.

\textbf{Topic Evolution}\quad During memory search, exploration driven purely by local similarity may focus on a single semantic aspect of a query, which can be insufficient for complex questions involving multiple facets. To address this issue, we introduce a topic layer over the accumulated event set
$\mathcal{E}^{(t)} = \bigcup_{i=1}^{t} \mathcal{E}_i$.
Each topic $z_k \in \mathcal{Z}^{(t)}$ represents a semantic cluster of related events, and the topic--event associations are maintained in $\mathcal{A}^{(t)}$, indicating which events belong to each topic.
This topic layer provides a coarse-grained semantic organization of events, which complements the fine-grained logical structure defined by event relations and facilitates efficient multi-path exploration during memory search.

At the initial stage (e.g., $t=1$), when no topic structure exists, we use K-means to perform topic clustering over the extracted events to initialize the topic set:
\begin{equation}
\mathcal{Z}^{(1)} = \{z_1, z_2, \dots, z_k\} = \Phi_{\mathrm{clu}}(\mathcal{E}^{(1)}; k).
\end{equation}

As memory grows over time, we update topic--event associations in an online manner.
For each newly integrated event $e_{\text{new}} \in \mathcal{E}_{t+1}$, we identify its most similar topic from the existing topic set $\mathcal{Z}^{(t)}$ based on semantic similarity.
If the similarity exceeds a threshold $\delta$, the event is assigned to that topic; otherwise, a new topic node is created to capture a previously unseen semantic direction.
This process incrementally updates both the topic set $\mathcal{Z}^{(t)}$ and the topic--event associations $\mathcal{A}^{(t)}$.

To prevent semantic drift introduced by incremental updates, we periodically re-cluster all accumulated events:
\begin{equation}
\mathcal{Z}^{(t+1)} \leftarrow \Phi_{\mathrm{clu}}(\mathcal{E}^{(t+1)}; k)
\quad \text{when } t \bmod T = 0.
\notag
\end{equation}
This strategy balances stability during online updates with global coherence over memory growth.

By treating temporally situated events as primary memory units, the resulting event graph preserves narrative structure and event-level semantics, which are often lost in triple-based representations~\citep{yang-etal-2025-eventrag}.
From this perspective, memory itself serves as an explicit \emph{logic map} that guides subsequent search and reasoning. Prompts for all memory construction processes are provided in the Appendix~\ref{append:construction-prompt}.

\begin{table*}[t]
\centering
\resizebox{\textwidth}{!}{
\begin{tabular}{ll cccccccc cc}
\toprule
\multirow{2.5}{*}{\textbf{Model}} & \multirow{2.5}{*}{\textbf{Method}} 
& \multicolumn{2}{c}{\textbf{Single-hop}} 
& \multicolumn{2}{c}{\textbf{Multi-hop}} 
& \multicolumn{2}{c}{\textbf{Open-domain}} 
& \multicolumn{2}{c}{\textbf{Temporal}} 
& \multicolumn{2}{c}{\textbf{Average}} \\
\cmidrule(lr){3-4} \cmidrule(lr){5-6} \cmidrule(lr){7-8} \cmidrule(lr){9-10} \cmidrule(lr){11-12}

& & \textbf{F1} & \textbf{BLEU} 
& \textbf{F1} & \textbf{BLEU} 
& \textbf{F1} & \textbf{BLEU} 
& \textbf{F1} & \textbf{BLEU} 
& \textbf{F1} & \textbf{BLEU} \\
\midrule

\multirow{9}{*}{\rotatebox{90}{\textbf{GPT-4o-mini}}}
& \textit{Non-Graph-based} & & & & & & & & & & \\
& RAG & 52.19 & 46.80 & 32.17 & 23.59 & 23.21 & 18.88 & 30.77 & 25.99 & 42.25 & 36.47 \\
& Mem0 & 47.65 & 38.72 & \underline{38.72} & \underline{27.13} & \textbf{28.64} & \underline{21.58} & \underline{48.93} & \underline{40.51} & 45.10 & 35.90 \\
& MemoryOS & 48.62 & 42.99 & 35.27 & 25.22 & 20.02 & 15.52 & 41.15 & 30.76 & 42.84 & 35.47 \\
\cmidrule(lr){2-12}
& \textit{Graph-based} & & & & & & & & & & \\
& HippoRAG & \underline{54.84} & \underline{48.84} & 33.59 & 25.46 & \underline{28.59} & \textbf{23.89} & 48.17 & 39.32 & \underline{47.92} & \underline{41.02} \\
& A-Mem & 44.65 & 37.06 & 27.02 & 20.09 & 12.14 & 12.00 & 45.85 & 36.67 & 39.65 & 32.31 \\
& CAM & 50.58 & 44.36 & 33.55 & 24.18 & 18.23 & 12.77 & 44.14 & 38.28 & 44.10 & 37.43 \\
& \cellcolor[RGB]{235,245,250}\ours{} & \cellcolor[RGB]{235,245,250}\textbf{57.36} & \cellcolor[RGB]{235,245,250}\textbf{49.79} & \cellcolor[RGB]{235,245,250}\textbf{38.84} & \cellcolor[RGB]{235,245,250}\textbf{27.98} & \cellcolor[RGB]{235,245,250}26.61 & \cellcolor[RGB]{235,245,250}20.01 & \cellcolor[RGB]{235,245,250}\textbf{57.96} & \cellcolor[RGB]{235,245,250}\textbf{50.51} & \cellcolor[RGB]{235,245,250}\textbf{52.18} & \cellcolor[RGB]{235,245,250}\textbf{44.09} \\
\midrule

\multirow{9}{*}{\rotatebox{90}{\textbf{Qwen2.5-14B}}}
& \textit{Non-Graph-based} & & & & & & & & & & \\
& RAG & 49.79 & 43.95 & 28.11 & 21.43 & 20.42 & 17.40 & 24.73 & 20.02 & 38.77 & 33.18 \\
& Mem0 & 42.58 & 35.15 & 31.73 & 24.82 & 15.03 & 11.28 & 28.96 & 26.24 & 36.04 & 29.91 \\
& MemoryOS & 46.33 & 41.62 & \underline{38.19} & \underline{29.26} & 20.27 & 15.94 & 32.24 & 27.86 & 40.28 & 34.89 \\
\cmidrule(lr){2-12}
& \textit{Graph-based} & & & & & & & & & & \\
& HippoRAG & 42.45 & 37.14 & 27.57 & 20.62 & 19.74 & 15.81 & 30.66 & 26.33 & 35.85 & 30.53 \\
& A-Mem & 33.75 & 30.04 & 22.09 & 15.28 & 13.49 & 10.74 & 27.19 & 22.05 & 28.98 & 24.47 \\
& CAM & \underline{50.39} & \underline{45.59} & 34.50 & 24.62 & \underline{23.86} & \underline{20.84} & \underline{44.70} & \underline{36.30} & \underline{44.64} & \underline{38.27} \\
& \cellcolor[RGB]{235,245,250}\ours{} & \cellcolor[RGB]{235,245,250}\textbf{61.02} & \cellcolor[RGB]{235,245,250}\textbf{55.93} & \cellcolor[RGB]{235,245,250}\textbf{42.32} & \cellcolor[RGB]{235,245,250}\textbf{32.66} & \cellcolor[RGB]{235,245,250}\textbf{25.88} & \cellcolor[RGB]{235,245,250}\textbf{22.01} & \cellcolor[RGB]{235,245,250}\textbf{47.18} & \cellcolor[RGB]{235,245,250}\textbf{39.69} & \cellcolor[RGB]{235,245,250}\textbf{52.52} & \cellcolor[RGB]{235,245,250}\textbf{46.17} \\
\bottomrule
\end{tabular}}
\caption{
Performance comparison on the LoCoMo benchmark, covering single-hop, multi-hop, open-domain, and temporal settings. We report F1 and BLEU-1 scores (\%). Best results are highlighted in \textbf{bold}, and second-best results are \underline{underlined}.
}
\label{tab:main}
\end{table*}

\begin{table}[t]
  \centering
  \small
  \begin{tabular}{l l c c}
    \toprule
    Model & Method & F1 & BLEU \\
    \midrule
    \multirow{7}{*}{GPT-4o-mini} 
      & RAG        & 28.99 & 25.68 \\
      & Mem0       & 29.98 & 23.34 \\
      & MemoryOS   & 25.58 & 21.74 \\
      & HippoRAG   & 28.77 & 23.04 \\
      & A-Mem      & 27.01 & 23.17 \\
      & CAM        & \underline{33.55} & \underline{29.74} \\
      & \cellcolor[RGB]{235,245,250}\ours{} & \cellcolor[RGB]{235,245,250}\textbf{39.04} & \cellcolor[RGB]{235,245,250}\textbf{35.23} \\
    \midrule
    \multirow{7}{*}{Qwen2.5-14B}
      & RAG        & 25.82 & 20.65 \\
      & Mem0       & 26.94 & 22.01 \\
      & MemoryOS   & 22.17 & 19.32 \\
      & HippoRAG   & 22.10 & 17.77 \\
      & A-Mem      & 25.37 & 20.94 \\
      & CAM        & \underline{27.87} & \underline{23.47} \\
      & \cellcolor[RGB]{235,245,250}\ours{} & \cellcolor[RGB]{235,245,250}\textbf{35.90} & \cellcolor[RGB]{235,245,250}\textbf{28.66} \\
    \bottomrule
  \end{tabular}
\caption{Results on 298 questions belonging to 10 documents randomly sampled from the NarrativeQA. We do the sample since the full test set contains over 10,000 questions and is prohibitively large for long-context evaluation.}
  \label{tab:narrativeQA}
\end{table}

\subsection{Active Multi-Path Memory Search}
\label{ssec:method_search}

With the event graph constructed as a structured logic map, memory search proceeds through active navigation and reasoning.
Given a query $q$, the goal is to retrieve a small set of event nodes that provide sufficient evidence.

\ours{} adopts a principle of guided active evidence construction. Reasoning is performed through traversal of the event graph, while only distilled evidence is passed to the final answer model.
To support this process, we implement memory search $\Psi$ using three LLM-based agents: a \textbf{Planner}, multiple \textbf{Explorers}, and an \textbf{Responder}.
Prompts for all agents are provided in the Appendix~\ref{append:search-prompt}.

\subsubsection{Planner}

Given a query $q$, the \textbf{Planner} decomposes it into a small set of subgoals,
\begin{equation}
\mathcal{H}_q = \Psi_{\mathrm{plan}}(q), \qquad |\mathcal{H}_q| \in [2,5],
\end{equation}
where each subgoal captures a distinct aspect that the search should cover. The Planner maintains a binary satisfaction vector $\mathbf{s} \in \{0,1\}^K$ to indicate which subgoals have been supported by the currently collected evidence.
This explicit progress signal provides a notion of the search stage and guides exploration toward unsatisfied subgoals.

If the search fails to terminate with sufficient evidence in the current round, the Planner performs gap-aware refinement.
It generates a refined query by conditioning on the current query, the collected evidence, and the remaining unsatisfied subgoals,
\begin{equation}
q^{(r+1)} = \Psi_{\mathrm{ref}}(q^{(r)}, \mathcal{H}_q, \mathbf{s}),
\end{equation}
where refinement focuses on unfinished subgoals.
This design yields a closed-loop process that alternates between exploration and query refinement.

\subsubsection{Explorer}

Active searching and reasoning over the memory is carried out by a set of \textbf{Explorer} agents.
Guided by the memory topology structure, each Explorer operates directly on the event graph and decides which nodes to retain as evidence and how exploration should proceed.

\noindent\textbf{Localization}\quad
Before graph traversal, exploration is first localized to determine where to begin.
Candidate event nodes are retrieved by ranking their embedding similarity to the query, and the top-$k$ results are selected.
Since these events are often highly similar and may focus on a single aspect, the Explorer further selects candidates from the first $p$ distinct topic clusters appearing in the ranked list.
From the resulting candidate set $\mathcal{C}_q$, the starting nodes are selected as:
\begin{equation}
\mathcal{S}_q = \Psi_{\mathrm{start}}(q, \mathcal{C}_q),
\notag
\end{equation}
where $\Psi_{\mathrm{start}}$ denotes an LLM-based selection operator.
The selected starting nodes are then inserted into a globally maintained queue to initialize subsequent exploration.

\noindent\textbf{Navigation}\quad
Guided by the event-graph topology, exploration proceeds step by step.
At each visited event node $e$, an Explorer conditions on the query, the current subgoal status, the retained evidence, and the local graph context, including neighboring nodes.
Based on this information, the Explorer chooses an action from the action space $\{\textsc{Skip}, \textsc{Expand}, \textsc{Answer}\}$:
\begin{equation}
a = \Psi_{\mathrm{cho}}(q, e, \hat{\mathcal{E}}, \mathcal{N}(e), \mathbf{s}),
\end{equation}
where $\hat{\mathcal{E}}$ denotes the current evidence set and $\mathcal{N}(e)$ denotes neighboring events with typed relations.
\textsc{Skip} discards the current node, \textsc{Expand} retains it as evidence and continues exploration, and \textsc{Answer} terminates the current path when sufficient evidence has been collected.
When \textsc{Expand} is selected, the evidence set is updated as:
\begin{equation}
\hat{\mathcal{E}}^{(t+1)} =
\begin{cases}
\hat{\mathcal{E}}^{(t)}, & \text{if } a =\text{SKIP}, \\
\hat{\mathcal{E}}^{(t)} \cup \{e\}, & \text{otherwise}. \\
\end{cases}
\end{equation}
Each retained node is annotated with the subgoals it supports, enabling explicit progress tracking.

This decision process operationalizes our key insight that \emph{topology carries logic}:
relations constrain exploration paths and guide reasoning over structured dependencies, rather than flat and isolated text.

\noindent\textbf{Coordination}\quad
Multiple Explorers run in parallel, each initialized from a different starting node.
They share a global state that records visited nodes, retained evidence, and subgoal progress.
All candidate nodes encountered during traversal are scheduled through a single global priority queue. The priority of a candidate node $u$ is defined by its embedding similarity to unsatisfied subgoals:
\begin{equation}
p(u)=\max_{j:s_j=0}\ \mathrm{sim}\!\big(v(s_u), v(h_j)\big),
\end{equation}
where $s_u$ denotes the summary of $u$ and $h_j$ denotes a subgoal.
This subgoal-driven scheduling reduces redundant exploration and promotes complementary coverage across paths, enabling efficient multi-path reasoning over the event graph.

\begin{figure*}[t]
  \centering
  \includegraphics[width=\textwidth]{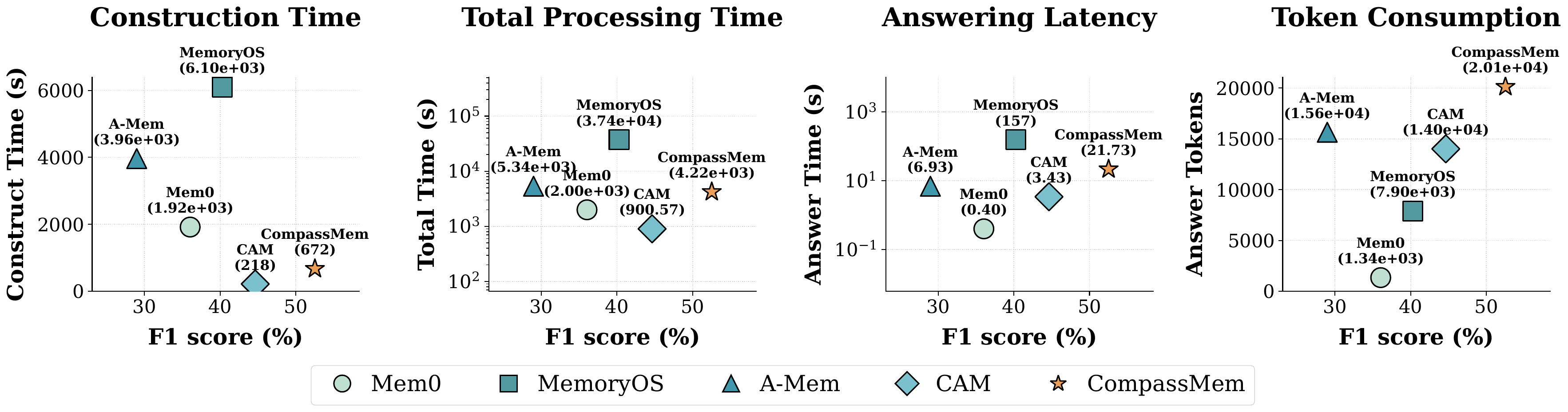}
  \caption{Efficiency–performance trade-off across memory frameworks. Scatter plots compare F1 with construction time, total processing time, per-question latency, and token consumption.}
  \label{fig:efficiency}
\end{figure*}

\subsubsection{Responder}
The \textbf{Responder} is invoked when the global candidate queue becomes empty, and all subgoals are satisfied.
If the queue becomes empty while some subgoals remain unsatisfied, the system returns to the Planner to start the second round search.

Upon termination, the search returns a concise evidence set $\mathcal{M}^{(t)}\!\mid_q = \hat{\mathcal{E}}$.
If no evidence is retained, we fall back to the initial top-$k$ retrieved candidates.
The Responder then generates the final output, ensuring that generation conditions only on distilled evidence while reasoning is carried out through structured navigation on the logic map.

\section{Experiment}

\subsection{Experimental Settings}

\noindent\textbf{Benchmarks}\quad
We evaluate \ours{} on two long-context reasoning benchmarks, \textbf{LoCoMo} and \textbf{NarrativeQA}. LoCoMo focuses on conversational question answering, while NarrativeQA targets narrative understanding. Detailed dataset descriptions are provided in the Appendix~\ref{append:dataset}.

\noindent\textbf{Backbone Models}\quad
We use \textbf{GPT-4o-mini} as a closed-source model, and \textbf{Qwen2.5-14B-Instruct} as an open-source model. Qwen is deployed with vLLM, while GPT is accessed via API.
All methods use \textbf{BGE-M3} for all mentioned embeddings.

\noindent\textbf{Baselines}\quad
We compare \ours{} with non-graph baselines, including RAG, Mem0~\citep{mem0prateek}, and MemoryOS~\citep{memoryoskang}, as well as graph-based baselines such as HippoRAG~\citep{hipporag}, A-Mem~\citep{amemxu}, and CAM~\citep{camli}. Official implementations or reported settings are used when available, with full implementation details provided in the Appendix~\ref{append:baseline_imply}.

\subsection{Main Results}

We now present the main experimental results and several key observations. Addational results are provided in the Appendix~\ref{sec:detailed_stats}.

(1) Table~\ref{tab:main} reports results on LoCoMo across question types. While most methods handle single-hop questions reasonably well, performance drops sharply on multi-hop and temporal QA. In contrast, \ours{} consistently achieves the strongest results. On GPT-4o-mini, it improves average F1 from 47.92\% (HippoRAG) to 52.18\%, with a large gain on temporal questions (57.96\% vs.\ 48.93\%). On Qwen2.5-14B, \ours{} further reaches 52.52\%  F1 and achieves the best performance on all subsets. These results demonstrate the benefit of event-graph memory with logic-aware retrieval for reasoning-intensive QA.

(2) Table~\ref{tab:narrativeQA} presents results on NarrativeQA, which requires long-range narrative understanding and evidence aggregation. \ours{} consistently outperforms all baselines, surpassing the strongest competitor CAM by over 5\% F1 on GPT-4o-mini and more than 8\% F1 on Qwen2.5-14B. This demonstrates the effectiveness of event-centric memory with explicit relations for retrieving globally relevant evidence in long narratives.

(3) Across both benchmarks, \ours{} shows consistent and robust improvements. Notably, the strongest baselines are generally graph-based, supporting the importance of structured memory. \ours{} further advances these methods by modeling memory at the event level with logic-aware relations, yielding the largest gains on tasks that require complex retrieval and reasoning.

\begin{figure}[t]
  \centering
  \includegraphics[width=\columnwidth]{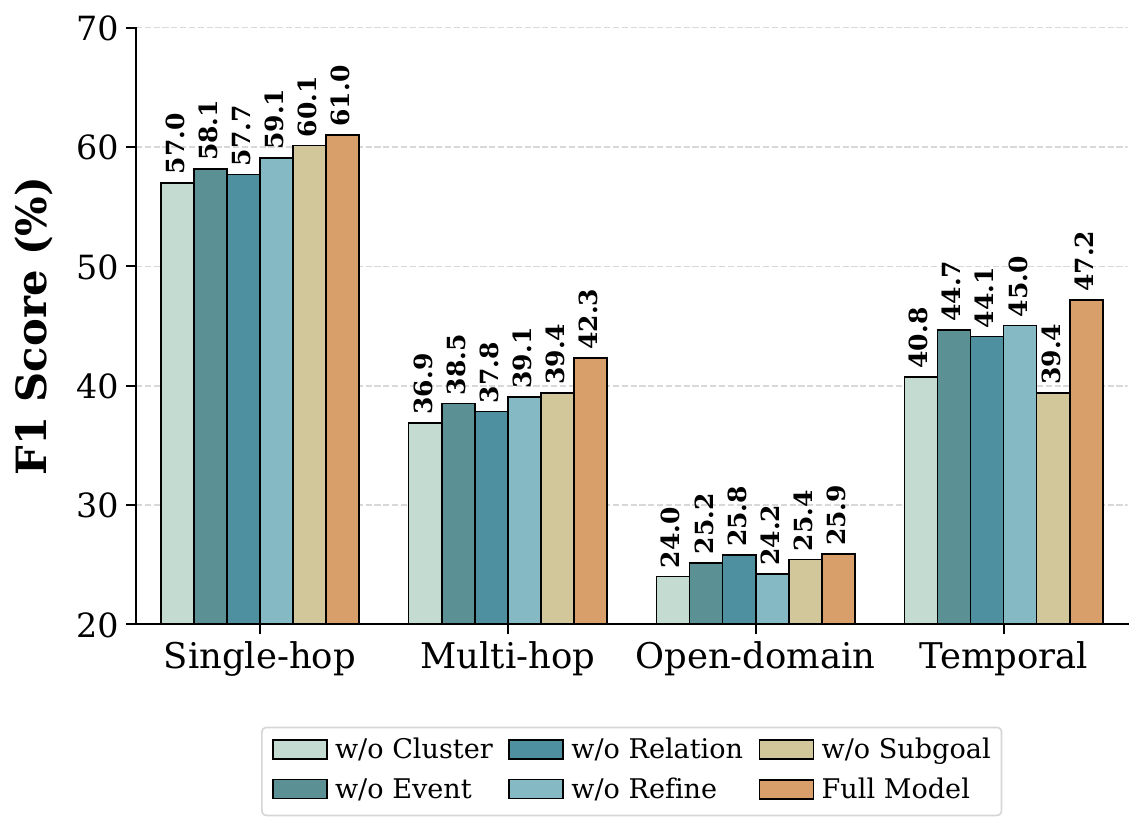}
  \caption{Ablation results on LoCoMo.}
  \label{fig:ablation}
\end{figure}

\begin{figure}[t]
  \includegraphics[width=\columnwidth]{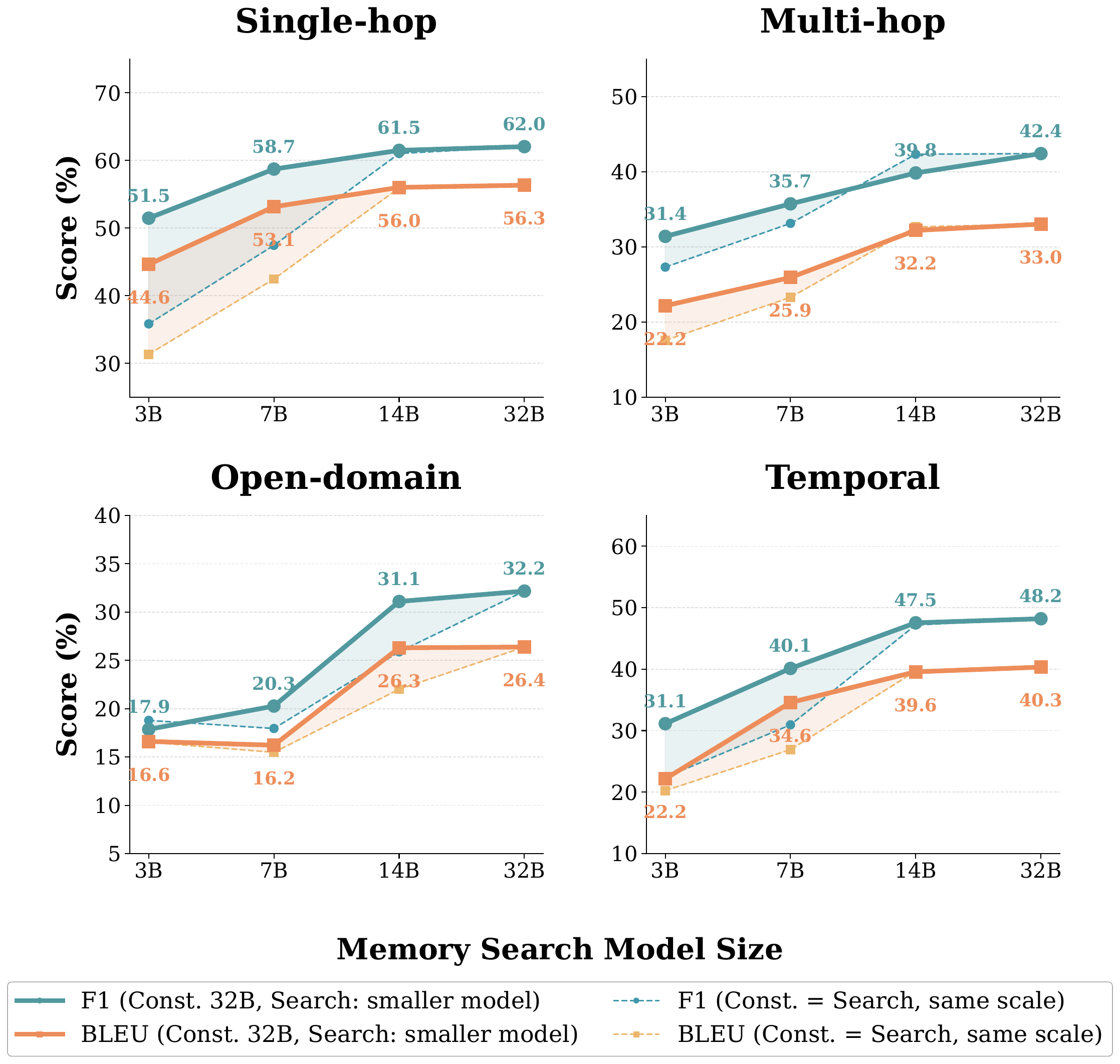}
  \caption{Scaling results comparing fixed high-capacity and scale-consistent memory construction. Shaded areas show gains from stronger construction models.}
  \label{fig:scaling}
\end{figure}

(4) We further analyze efficiency on a representative LoCoMo conversation set, ramdomly selected from ten sessions with identical settings, as shown in Figure~\ref{fig:efficiency}. \ours{} achieves low memory construction time cost, substantially lower than Mem0, A-Mem, and MemoryOS. Our total processing time and per-question latency are comparable to Mem0 and A-Mem, and markedly lower than MemoryOS. Although \ours{} uses more tokens, this cost is accompanied by substantial performance gains. Overall, \ours{} delivers strong reasoning improvements while maintaining practical computational efficiency.

\begin{figure}[t]
  \includegraphics[width=\columnwidth]{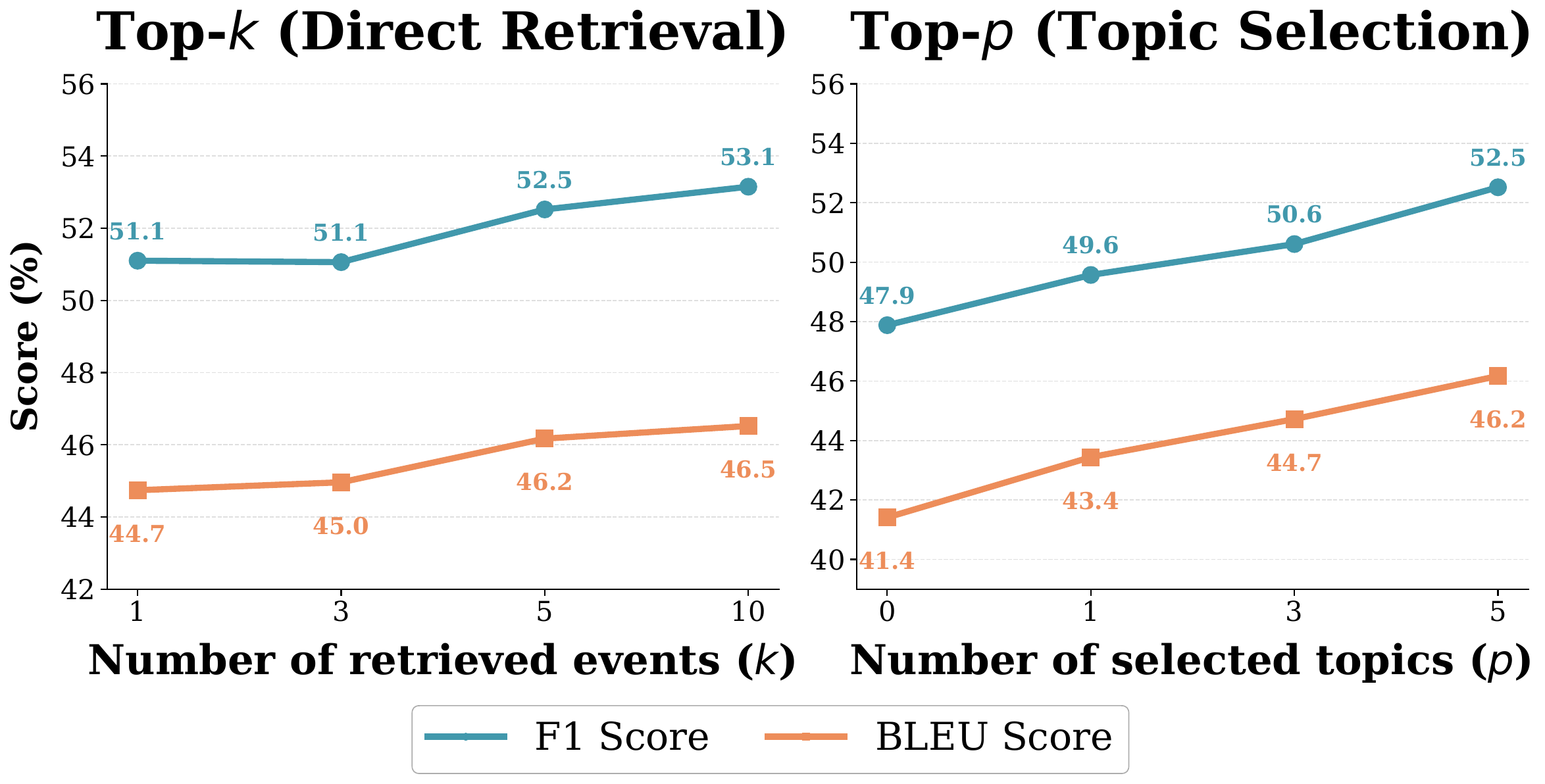}
  \caption{Sensitivity analysis of \ours{} with respect to localization hyperparameters}
  \label{fig:sensitivity}
\end{figure}

\subsection{Further Analysis}
\label{ssec:futher_ana}
We further conduct in-depth analyses to better understand the behavior of \ours{}. 

\noindent\textbf{Ablation Study}\quad
To examine the effectiveness of individual components in \ours{}, we conduct an ablation study by systematically removing key modules. Specifically, we evaluate variants that (i) remove topic clustering, (ii) replace event units with fixed-length chunks to eliminate event modeling, where the chunk length is set to the average size of extracted events (approximately 100 tokens), (iii) remove edges to discard explicit relations, (iv) disable query refinement to prevent second-round exploration, and (v) remove subgoal generation.
Figure~\ref{fig:ablation} reports the ablation results across question categories. Removing any component leads to consistent performance drops, confirming the contribution of each module. In particular, multi-hop and temporal questions are most affected, while single-hop and open-domain questions show smaller degradation due to lower reasoning complexity.

\noindent\textbf{Impact of Model Size}\quad
We examine the scalability of \ours{}. \Cref{fig:scaling} shows that \ours{} continues to improve as model scale increases when the same backbone is used for both memory construction and search. We further evaluate a decoupled setting where memory is constructed with a high-capacity model (Qwen2.5-32B) while search and response generation use smaller models. This configuration yields clear improvements over scale-matched baselines. These results suggest that high-quality memory structures built offline can effectively support downstream reasoning, even when paired with lightweight search models.

\noindent\textbf{Impact of Location Hyperparameters}\quad
\Cref{fig:sensitivity} analyzes the sensitivity of \ours{} to two localization hyperparameters: the direct retrieval size $k$, and the topic selection size $p$. Overall, introducing topic-based selection ($p>0$) consistently improves performance compared to the no-clustering setting, and larger values of $p$ lead to steadily better results. This suggests that selecting starting nodes from multiple semantic topics helps diversify exploration and reduces bias toward a single semantic view. Similarly, increasing the retrieval size $k$ provides a broader pool of candidate events and yields monotonic performance gains, indicating that richer initial retrieval better supports downstream search.

\noindent\textbf{Impact of Model Thinking Ability}\quad
Table~\ref{tab:qwen3} reports LoCoMo results on Qwen3-8B, a backbone equipped with explicit thinking capability. All methods benefit from the stronger reasoning capacity, with noticeable improvements on multi-hop and temporal questions compared to non-thinking models. Nevertheless, \ours{} consistently achieves the best performance across all task categories. The gains indicate that explicit reasoning alone is insufficient. Effective memory organization and logic-aware retrieval remain critical for fully exploiting the backbone’s thinking ability.
\begin{table}[t] 
\centering
\resizebox{\columnwidth}{!}{ 
\begin{tabular}{lcccccccc}
\toprule
\multirow{2}{*}{\textbf{Method}} & \multicolumn{2}{c}{\textbf{Single-hop}} & \multicolumn{2}{c}{\textbf{Multi-hop}} & \multicolumn{2}{c}{\textbf{Open-domain}} & \multicolumn{2}{c}{\textbf{Temporal}} \\ \cmidrule(lr){2-3} \cmidrule(lr){4-5} \cmidrule(lr){6-7} \cmidrule(lr){8-9}
 & F1 & BLEU & F1 & BLEU & F1 & BLEU & F1 & BLEU \\ \midrule
RAG & 44.26 & 37.98 & 23.84 & 15.46 & 11.39 & 8.33 & 17.88 & 13.09 \\
MemO & 37.50 & 31.76 & 23.58 & 15.17 & 14.37 & 11.48 & 41.29 & 30.37 \\
MemoryOS & 40.87 & 35.84 & 24.71 & 19.28 & 16.09 & 14.50 & 39.41 & 28.71 \\
HippoRAG & \underline{46.12} & \underline{40.58} & 31.62 & 24.52 & \underline{22.04} & \underline{17.47} & 40.39 & 32.94 \\
A-Mem & 44.57 & 39.37 & 28.53 & 20.16 & 18.35 & 15.23 & 31.60 & 23.49 \\
CAM & 45.79 & 38.48 & \underline{34.07} & \underline{26.01} & 19.96 & 16.36 & \underline{43.82} & \underline{36.11} \\ \midrule
\textbf{Ours} & \textbf{50.04} & \textbf{43.40} & \textbf{35.33} & \textbf{27.86} & \textbf{28.02} & \textbf{23.25} & \textbf{49.35} & \textbf{38.91} \\ \bottomrule
\end{tabular}
}
\caption{Results of LoCoMo on Qwen3-8B.}
\label{tab:qwen3}
\end{table}

\section{Conclusion}
We presented \ours{}, an event-centric memory framework that rethinks agent memory as a structured logic map rather than a flat storage. By organizing experiences into coherent events and explicitly modeling their logical relations, \ours{} enables memory to actively guide searching and reasoning. Experiments on dialogue and long-document demonstrate that this design provides strong and consistent benefits, particularly for reasoning-intensive tasks. We hope this work encourages future research on memory structures that more directly support long-horizon reasoning and decision-making in intelligent agents.

\section*{Limitations}
While \ours{} shows consistent gains, it has several limitations. 

First, the quality of the Event Graph depends on event segmentation and relation extraction. In this work, we adopt a naive LLM-based pipeline; more fine-grained and robust segmentation may further improve memory quality, and we leave this direction for future work. 

 Second, our evaluation focuses on a set of representative benchmarks. Demonstrating the effectiveness of \ours{} across a broader range of tasks and agent settings would further strengthen its applicability.

\section*{Ethical considerations}
This work studies agent memory architectures for long-context reasoning and does not introduce new datasets. All experiments are conducted on publicly available benchmarks, LoCoMo and NarrativeQA, which do not contain sensitive personal information. We do not intentionally collect, infer, or generate content that identifies specific individuals.

\bibliography{custom}

@article{mem0prateek,
  author       = {Prateek Chhikara and
                  Dev Khant and
                  Saket Aryan and
                  Taranjeet Singh and
                  Deshraj Yadav},
  title        = {Mem0: Building Production-Ready {AI} Agents with Scalable Long-Term
                  Memory},
  journal      = {CoRR},
  volume       = {abs/2504.19413},
  year         = {2025}
}

@inproceedings{memorybankzhong,
  author       = {Wanjun Zhong and
                  Lianghong Guo and
                  Qiqi Gao and
                  He Ye and
                  Yanlin Wang},
  title        = {MemoryBank: Enhancing Large Language Models with Long-Term Memory},
  booktitle    = {{AAAI}},
  pages        = {19724--19731},
  publisher    = {{AAAI} Press},
  year         = {2024}
}

@article{memgptcharles,
  author       = {Charles Packer and
                  Vivian Fang and
                  Shishir G. Patil and
                  Kevin Lin and
                  Sarah Wooders and
                  Joseph E. Gonzalez},
  title        = {MemGPT: Towards LLMs as Operating Systems},
  journal      = {CoRR},
  volume       = {abs/2310.08560},
  year         = {2023}
}

@inproceedings{memtreealireza,
  author       = {Alireza Rezazadeh and
                  Zichao Li and
                  Wei Wei and
                  Yujia Bao},
  title        = {From Isolated Conversations to Hierarchical Schemas: Dynamic Tree
                  Memory Representation for LLMs},
  booktitle    = {{ICLR}},
  publisher    = {OpenReview.net},
  year         = {2025}
}

@article{memosli,
  author       = {Zhiyu Li and
                  Shichao Song and
                  Chenyang Xi and
                  Hanyu Wang and
                  Chen Tang and
                  Simin Niu and
                  Ding Chen and
                  Jiawei Yang and
                  Chunyu Li and
                  Qingchen Yu and
                  Jihao Zhao and
                  Yezhaohui Wang and
                  Peng Liu and
                  Zehao Lin and
                  Pengyuan Wang and
                  Jiahao Huo and
                  Tianyi Chen and
                  Kai Chen and
                  Kehang Li and
                  Zhen Tao and
                  Junpeng Ren and
                  Huayi Lai and
                  Hao Wu and
                  Bo Tang and
                  Zhenren Wang and
                  Zhaoxin Fan and
                  Ningyu Zhang and
                  Linfeng Zhang and
                  Junchi Yan and
                  Mingchuan Yang and
                  Tong Xu and
                  Wei Xu and
                  Huajun Chen and
                  Haofeng Wang and
                  Hongkang Yang and
                  Wentao Zhang and
                  Zhi{-}Qin John Xu and
                  Siheng Chen and
                  Feiyu Xiong},
  title        = {MemOS: {A} Memory {OS} for {AI} System},
  journal      = {CoRR},
  volume       = {abs/2507.03724},
  year         = {2025}
}

@article{memoryoskang,
  author       = {Jiazheng Kang and
                  Mingming Ji and
                  Zhe Zhao and
                  Ting Bai},
  title        = {Memory {OS} of {AI} Agent},
  journal      = {CoRR},
  volume       = {abs/2506.06326},
  year         = {2025}
}

@article{zeprasmussen,
  author       = {Preston Rasmussen and
                  Pavlo Paliychuk and
                  Travis Beauvais and
                  Jack Ryan and
                  Daniel Chalef},
  title        = {Zep: {A} Temporal Knowledge Graph Architecture for Agent Memory},
  journal      = {CoRR},
  volume       = {abs/2501.13956},
  year         = {2025}
}

@article{gmemoryzhang,
  author       = {Guibin Zhang and
                  Muxin Fu and
                  Guancheng Wan and
                  Miao Yu and
                  Kun Wang and
                  Shuicheng Yan},
  title        = {G-Memory: Tracing Hierarchical Memory for Multi-Agent Systems},
  journal      = {CoRR},
  volume       = {abs/2506.07398},
  year         = {2025}
}

@article{sgmemwu,
  author       = {Yaxiong Wu and
                  Yongyue Zhang and
                  Sheng Liang and
                  Yong Liu},
  title        = {SGMem: Sentence Graph Memory for Long-Term Conversational Agents},
  journal      = {CoRR},
  volume       = {abs/2509.21212},
  year         = {2025}
}

@article{memoryr1yan,
  author       = {Sikuan Yan and
                  Xiufeng Yang and
                  Zuchao Huang and
                  Ercong Nie and
                  Zifeng Ding and
                  Zonggen Li and
                  Xiaowen Ma and
                  Hinrich Sch{\"{u}}tze and
                  Volker Tresp and
                  Yunpu Ma},
  title        = {Memory-R1: Enhancing Large Language Model Agents to Manage and Utilize
                  Memories via Reinforcement Learning},
  journal      = {CoRR},
  volume       = {abs/2508.19828},
  year         = {2025}
}

@article{memalphawang,
  author       = {Yu Wang and
                  Ryuichi Takanobu and
                  Zhiqi Liang and
                  Yuzhen Mao and
                  Yuanzhe Hu and
                  Julian J. McAuley and
                  Xiaojian Wu},
  title        = {Mem-{\(\alpha\)}: Learning Memory Construction via Reinforcement Learning},
  journal      = {CoRR},
  volume       = {abs/2509.25911},
  year         = {2025}
}

@article{camli,
  author       = {Rui Li and
                  Zeyu Zhang and
                  Xiaohe Bo and
                  Zihang Tian and
                  Xu Chen and
                  Quanyu Dai and
                  Zhenhua Dong and
                  Ruiming Tang},
  title        = {{CAM:} {A} Constructivist View of Agentic Memory for LLM-Based Reading
                  Comprehension},
  journal      = {CoRR},
  volume       = {abs/2510.05520},
  year         = {2025}
}

@article{amemxu,
  author       = {Wujiang Xu and
                  Zujie Liang and
                  Kai Mei and
                  Hang Gao and
                  Juntao Tan and
                  Yongfeng Zhang},
  title        = {{A-MEM:} Agentic Memory for {LLM} Agents},
  journal      = {CoRR},
  volume       = {abs/2502.12110},
  year         = {2025}
}

@article{reasoningbankouyang,
  author       = {Siru Ouyang and
                  Jun Yan and
                  I{-}Hung Hsu and
                  Yanfei Chen and
                  Ke Jiang and
                  Zifeng Wang and
                  Rujun Han and
                  Long T. Le and
                  Samira Daruki and
                  Xiangru Tang and
                  Vishy Tirumalashetty and
                  George Lee and
                  Mahsan Rofouei and
                  Hangfei Lin and
                  Jiawei Han and
                  Chen{-}Yu Lee and
                  Tomas Pfister},
  title        = {ReasoningBank: Scaling Agent Self-Evolving with Reasoning Memory},
  journal      = {CoRR},
  volume       = {abs/2509.25140},
  year         = {2025}
}

@article{agentmemorysurveyzhang,
  author       = {Zeyu Zhang and
                  Quanyu Dai and
                  Xiaohe Bo and
                  Chen Ma and
                  Rui Li and
                  Xu Chen and
                  Jieming Zhu and
                  Zhenhua Dong and
                  Ji{-}Rong Wen},
  title        = {A Survey on the Memory Mechanism of Large Language Model-based Agents},
  journal      = {{ACM} Trans. Inf. Syst.},
  volume       = {43},
  number       = {6},
  pages        = {155:1--155:47},
  year         = {2025}
}

@article{baldassano2017discovering,
  title={Discovering event structure in continuous narrative perception and memory},
  author={Baldassano, Christopher and Chen, Janice and Zadbood, Asieh and Pillow, Jonathan W and Hasson, Uri and Norman, Kenneth A},
  journal={Neuron},
  volume={95},
  number={3},
  pages={709--721},
  year={2017},
  publisher={Elsevier}
}

@misc{hu2025memoryageaiagents,
      title={Memory in the Age of AI Agents}, 
      author={Yuyang Hu and Shichun Liu and Yanwei Yue and Guibin Zhang and Boyang Liu and Fangyi Zhu and Jiahang Lin and Honglin Guo and Shihan Dou and Zhiheng Xi and Senjie Jin and Jiejun Tan and Yanbin Yin and Jiongnan Liu and Zeyu Zhang and Zhongxiang Sun and Yutao Zhu and Hao Sun and Boci Peng and Zhenrong Cheng and Xuanbo Fan and Jiaxin Guo and Xinlei Yu and Zhenhong Zhou and Zewen Hu and Jiahao Huo and Junhao Wang and Yuwei Niu and Yu Wang and Zhenfei Yin and Xiaobin Hu and Yue Liao and Qiankun Li and Kun Wang and Wangchunshu Zhou and Yixin Liu and Dawei Cheng and Qi Zhang and Tao Gui and Shirui Pan and Yan Zhang and Philip Torr and Zhicheng Dou and Ji-Rong Wen and Xuanjing Huang and Yu-Gang Jiang and Shuicheng Yan},
      year={2025},
      eprint={2512.13564},
      archivePrefix={arXiv},
      primaryClass={cs.CL},
}

@inproceedings{yang-etal-2025-eventrag,
    title = "{E}vent{RAG}: Enhancing {LLM} Generation with Event Knowledge Graphs",
    author = "Yang, Zairun  and
      Wang, Yilin  and
      Shi, Zhengyan  and
      Yao, Yuan  and
      Liang, Lei  and
      Ding, Keyan  and
      Yilmaz, Emine  and
      Chen, Huajun  and
      Zhang, Qiang",
    editor = "Che, Wanxiang  and
      Nabende, Joyce  and
      Shutova, Ekaterina  and
      Pilehvar, Mohammad Taher",
    booktitle = "Proceedings of the 63rd Annual Meeting of the Association for Computational Linguistics (Volume 1: Long Papers)",
    month = jul,
    year = "2025",
    address = "Vienna, Austria",
    publisher = "Association for Computational Linguistics",
    url = "https://aclanthology.org/2025.acl-long.830/",
    doi = "10.18653/v1/2025.acl-long.830",
    pages = "16967--16979",
    ISBN = "979-8-89176-251-0"
}

@inproceedings{hipporag,
  author       = {Bernal Jim{\'{e}}nez Guti{\'{e}}rrez and
                  Yiheng Shu and
                  Weijian Qi and
                  Sizhe Zhou and
                  Yu Su},
  title        = {From {RAG} to Memory: Non-Parametric Continual Learning for Large
                  Language Models},
  booktitle    = {{ICML}},
  publisher    = {OpenReview.net},
  year         = {2025}
}

@article{agentsurveywang,
  author       = {Lei Wang and
                  Chen Ma and
                  Xueyang Feng and
                  Zeyu Zhang and
                  Hao Yang and
                  Jingsen Zhang and
                  Zhiyuan Chen and
                  Jiakai Tang and
                  Xu Chen and
                  Yankai Lin and
                  Wayne Xin Zhao and
                  Zhewei Wei and
                  Jirong Wen},
  title        = {A survey on large language model based autonomous agents},
  journal      = {Frontiers Comput. Sci.},
  volume       = {18},
  number       = {6},
  pages        = {186345},
  year         = {2024}
}

@misc{zhang2025memevolvemetaevolutionagentmemory,
      title={MemEvolve: Meta-Evolution of Agent Memory Systems}, 
      author={Guibin Zhang and Haotian Ren and Chong Zhan and Zhenhong Zhou and Junhao Wang and He Zhu and Wangchunshu Zhou and Shuicheng Yan},
      year={2025},
      eprint={2512.18746},
      archivePrefix={arXiv},
      primaryClass={cs.CL},
      url={https://arxiv.org/abs/2512.18746}, 
}

@article{memgenzhang,
  author       = {Guibin Zhang and
                  Muxin Fu and
                  Shuicheng Yan},
  title        = {MemGen: Weaving Generative Latent Memory for Self-Evolving Agents},
  journal      = {CoRR},
  volume       = {abs/2509.24704},
  year         = {2025}
}

@article{memoryhuawei,
  author       = {Yaxiong Wu and
                  Sheng Liang and
                  Chen Zhang and
                  Yichao Wang and
                  Yongyue Zhang and
                  Huifeng Guo and
                  Ruiming Tang and
                  Yong Liu},
  title        = {From Human Memory to {AI} Memory: {A} Survey on Memory Mechanisms
                  in the Era of LLMs},
  journal      = {CoRR},
  volume       = {abs/2504.15965},
  year         = {2025}
}

@misc{sun2025hierarchicalmemoryhighefficiencylongterm,
      title={Hierarchical Memory for High-Efficiency Long-Term Reasoning in LLM Agents}, 
      author={Haoran Sun and Shaoning Zeng},
      year={2025},
      eprint={2507.22925},
      archivePrefix={arXiv},
      primaryClass={cs.CL},
}

@article{zacks2007event,
  title={Event perception: a mind-brain perspective.},
  author={Zacks, Jeffrey M and Speer, Nicole K and Swallow, Khena M and Braver, Todd S and Reynolds, Jeremy R},
  journal={Psychological bulletin},
  volume={133},
  number={2},
  pages={273},
  year={2007},
  publisher={American Psychological Association}
}

@article{ezzyat2011constitutes,
  title={What constitutes an episode in episodic memory?},
  author={Ezzyat, Youssef and Davachi, Lila},
  journal={Psychological science},
  volume={22},
  number={2},
  pages={243--252},
  year={2011},
  publisher={Sage Publications Sage CA: Los Angeles, CA}
}

@article{anderson1983spreading,
  title={A spreading activation theory of memory},
  author={Anderson, John R},
  journal={Journal of verbal learning and verbal behavior},
  volume={22},
  number={3},
  pages={261--295},
  year={1983},
  publisher={Elsevier}
}

@inproceedings{maharana-etal-2024-evaluating,
    title = "Evaluating Very Long-Term Conversational Memory of {LLM} Agents",
    author = "Maharana, Adyasha  and
      Lee, Dong-Ho  and
      Tulyakov, Sergey  and
      Bansal, Mohit  and
      Barbieri, Francesco  and
      Fang, Yuwei",
    editor = "Ku, Lun-Wei  and
      Martins, Andre  and
      Srikumar, Vivek",
    booktitle = "Proceedings of the 62nd Annual Meeting of the Association for Computational Linguistics (Volume 1: Long Papers)",
    month = aug,
    year = "2024",
    address = "Bangkok, Thailand",
    publisher = "Association for Computational Linguistics",
    url = "https://aclanthology.org/2024.acl-long.747/",
    doi = "10.18653/v1/2024.acl-long.747",
    pages = "13851--13870"
}

@misc{yan2025generalagenticmemorydeep,
      title={General Agentic Memory Via Deep Research}, 
      author={B. Y. Yan and Chaofan Li and Hongjin Qian and Shuqi Lu and Zheng Liu},
      year={2025},
      eprint={2511.18423},
      archivePrefix={arXiv},
      primaryClass={cs.CL},
      url={https://arxiv.org/abs/2511.18423}, 
}

@misc{kočiský2017narrativeqareadingcomprehensionchallenge,
      title={The NarrativeQA Reading Comprehension Challenge}, 
      author={Tomáš Kočiský and Jonathan Schwarz and Phil Blunsom and Chris Dyer and Karl Moritz Hermann and Gábor Melis and Edward Grefenstette},
      year={2017},
      eprint={1712.07040},
      archivePrefix={arXiv},
      primaryClass={cs.CL},
      url={https://arxiv.org/abs/1712.07040}, 
}

@article{dubrow2024event,
  title={Event and boundaries},
  author={DuBrow, Sarah and Kahana, MJ and Wagner, AD},
  journal={Oxford handbook of human memory},
  volume={1},
  year={2024},
  publisher={Oxford University Press Oxford}
}

\clearpage
\appendix
\section*{Appendix}
\label{sec:appendix}

\section{Event Segmentation Theory}
\label{append:EST}
Event Segmentation Theory (EST)~\citep{baldassano2017discovering,zacks2007event,ezzyat2011constitutes} is a framework in cognitive science and neuroscience that explains how humans parse continuous streams of perceptual experience into meaningful units, or events. According to this theory, when perceiving a dynamic environment, humans do not process information as an undifferentiated continuous flow. Instead, experience is automatically segmented into a sequence of relatively stable event episodes. Within each event, representations remain coherent and stable; when a salient change occurs, such as a shift in scene, action goals, or environmental state, an event boundary is triggered, prompting the construction of a new event representation model.

This segmentation process operates not only at the perceptual level but also plays a critical role in the encoding of event memories and their subsequent retrieval. Event Segmentation Theory emphasizes that human experience is not a continuous whole, but rather is composed of a series of identifiable event units. Such segmentation enhances perceptual efficiency and provides a fundamental basis for memory structuring and information retrieval~\citep{dubrow2024event}.

\section{Experiment Details}
\label{append:exp_imply}
\subsection{Dataset Descriptions}
\label{append:dataset}

\paragraph{LoCoMo} LoCoMo~\citep{maharana-etal-2024-evaluating} is a benchmark of very long-term conversational dialogues designed to evaluate long-range memory and reasoning capabilities in agent systems. The dataset consists of 10 extended conversations, each spanning dozens of sessions and hundreds of dialogue turns, with an average of around 600 turns and roughly 16K tokens per conversation. Questions in the LoCoMo QA evaluation are annotated with answer locations and categorized into types such as single-hop, multi-hop, open-domain, temporal reasoning, and adversarial, targeting different memory and inference challenges. In our experiments on LoCoMo QA, we follow standard practice in related work and do not use adversarial question data, which aligns with previous evaluations~\citep{mem0prateek, yan2025generalagenticmemorydeep,memoryoskang}.

\paragraph{NarrativeQA} NarrativeQA~\citep{kočiský2017narrativeqareadingcomprehensionchallenge} is a large-scale reading comprehension benchmark that assesses models’ ability to understand and reason over long narrative text such as books and movie scripts. The full NarrativeQA dataset contains on the order of tens of thousands of human-written question–answer pairs associated with over a thousand story documents, where questions require synthesis across global document structure rather than shallow pattern matching. Questions are constructed based on human-generated abstractive summaries, encouraging deep narrative understanding and integrative reasoning beyond local context overlaps. In our evaluation, we randomly sampled 10 long documents from the NarrativeQA corpus and used their associated 298 QA pairs to measure performance on long-range narrative question answering. This sampling strategy is adopted because the full NarrativeQA test set contains 10,557 questions, making exhaustive evaluation computationally prohibitive. The selected documents have an average length of around 60,000 tokens, which still poses a substantial challenge for long-context understanding and coherent evidence aggregation.

\subsection{CompassMem}
\label{append:compassmem_imply}
In \ours{}, we adopt a fixed set of hyperparameters across all main experiments.
During memory construction, newly extracted events are merged with existing ones when their semantic similarity exceeds a threshold of $0.9$, which helps reduce redundancy while preserving coherent event structure.
In the topic evolution stage, we apply the same similarity threshold ($0.9$) when merging events into existing topics, and perform periodic re-clustering every 4 construction steps to maintain semantic coherence over time.
For the LoCoMo benchmark, memory localization retrieves the top-$k{=}5$ candidate events based on embedding similarity. To encourage multi-perspective exploration, candidates are selected from the top-$p{=}5$ distinct topic clusters. Topic clustering is performed using $k$-means, where the number of clusters is automatically determined by the current memory size as
$n_{\text{clusters}} = \max(2, \min(\lfloor n_{\text{samples}} / 5 \rfloor, 50))$
During memory search, we employ three parallel Explorer agents to conduct multi-path traversal over the Event Graph. Query refinement is limited to a single additional round to control search complexity. Our choice of hyperparameters is motivated by the analysis in~\Cref{ssec:futher_ana}.

For NarrativeQA, where documents are substantially longer, we increase the retrieval scope to top-$k{=}10$ while keeping all other settings unchanged. This adjustment allows broader initial coverage without altering the overall search strategy.

\subsection{Baseline}
\label{append:baseline_imply}
\paragraph{Description} 
\begin{itemize}
    \item Mem0~\citep{mem0prateek}: A scalable long-term memory system that dynamically extracts, consolidates, and retrieves salient facts from ongoing dialogues or streams. It maintains a compact set of memory entries by continually updating and merging similar facts, avoiding redundancy, and retrieves only the most relevant facts rather than re-processing the full context.
    \item MemoryOS~\citep{memoryoskang}: A hierarchical memory architecture designed specifically for AI agents in long conversational interactions. It organizes memory into multiple tiers (short-, mid-, and long-term stores) and coordinates four core modules—memory storage, dynamic update, adaptive retrieval, and response generation—to maintain continuity, context coherence, and personalization over long dialogues.
    \item HippoRAG~\citep{hipporag}: A graph-based retrieval-augmented generation framework inspired by the hippocampal indexing theory of human long-term memory. It transforms documents into a knowledge graph and uses Personalized PageRank over concept seeds to integrate information across disparate contexts, enabling efficient single-step multi-hop retrieval. This structure allows deeper integration of new experiences and improved retrieval for reasoning-intensive tasks compared to standard RAG.
    \item A-Mem~\citep{amemxu}: An agentic memory system for LLM agents that dynamically organizes memory entries into an interconnected network using principles from human note-taking methods. When new memories are added, it generates structured notes with multiple attributes and connects them to related historical memories, enabling continuous memory evolution and contextual organization beyond fixed schemas.
    \item CAM~\citep{camli}: A structured memory framework grounded in constructivist theory, which organizes memory hierarchically and supports flexible integration and dynamic adaptation. It maintains overlapping clusters and hierarchical summaries and explores memory structure during retrieval in a way reminiscent of human associative processes, improving both performance and efficiency on long-text reading tasks.

\end{itemize}

\paragraph{Implementation}
For baselines that rely on chunk-based retrieval, we apply a unified preprocessing strategy by segmenting documents into fixed-length chunks of 512 tokens. For all such methods, we retrieve the top-$5$ most relevant chunks based on embedding similarity and use them as context for downstream reasoning or answer generation. This ensures a consistent retrieval budget across chunk-based baselines.

For memory-based baselines, we follow their original experimental settings and implementations as described in the corresponding papers or official codebases, without additional modification. This design ensures a fair comparison while preserving the intended behavior of each baseline.

\section{Detailed Search and Reasoning Statistics}
\label{sec:detailed_stats}

This section provides a detailed analysis of the search and reasoning behavior of \ours{} on the LoCoMo benchmark. We report aggregated statistics to characterize efficiency, exploration dynamics, and the role of planning and refinement during memory search.

\subsection{Overall Statistics}

Table~\ref{tab:overall_stats} summarizes the overall runtime, retrieval, and reasoning statistics across all 1,540 questions. On average, each query is processed within a moderate and stable time budget, indicating that active navigation over the Event Graph does not lead to excessive overhead. The median runtime is close to the mean, suggesting consistent behavior across different queries.

The Planner generates approximately three subgoals per question, providing structured guidance for exploration. While not all subgoals are fully satisfied, partial satisfaction is common, reflecting the varying availability of supporting evidence in memory. The high refinement rate indicates that iterative query adjustment plays an important role in addressing uncovered aspects during search.

\begin{table}[h!]
\centering
\small
\begin{tabular}{lr}
\toprule
\textbf{Total Questions} & 1540 \\
\midrule
\multicolumn{2}{l}{\textbf{Time Metrics}} \\
Total Time & 32136.5\,s \\
Avg. Time per Question & 20.87\,s \\
Median Time & 19.32\,s \\
Max Time & 65.38\,s \\
Min Time & 4.84\,s \\
\midrule
\multicolumn{2}{l}{\textbf{Subgoal Metrics}} \\
Avg. Subgoals & 3.04 \\
Avg. Subgoal Satisfaction & 68.3\% \\
Fully Satisfied & 594 (38.6\%) \\
\midrule
\multicolumn{2}{l}{\textbf{Retrieval Metrics}} \\
Avg. Retrieved Nodes & 50.0 \\
Avg. Initial Nodes & 3.7 \\
Avg. Similarity & 0.7374 \\
\midrule
\multicolumn{2}{l}{\textbf{Traversal Metrics}} \\
Avg. Paths & 2.5 \\
Avg. Total Steps & 7.5 \\
Avg. Path Length & 2.84 \\
Max Path Length & 11 \\
Avg. Max Rounds & 2.4 \\
\midrule
\multicolumn{2}{l}{\textbf{Action Distribution}} \\
Total Actions & 11595 \\
EXPAND & 7348 (63.4\%) \\
SKIP & 4230 (36.5\%) \\
ANSWER & 17 (0.1\%) \\
\midrule
\multicolumn{2}{l}{\textbf{Queue Metrics}} \\
Avg. Initial Queue Size & 3.7 \\
Avg. Max Queue Size & 3.7 \\
\midrule
\multicolumn{2}{l}{\textbf{Refinement Metrics}} \\
Refinement Count & 1176 \\
Refinement Rate & 76.4\% \\
\midrule
\multicolumn{2}{l}{\textbf{Kept Nodes}} \\
Avg. Kept Nodes & 3.15 \\
Max Kept Nodes & 14 \\
No Kept Nodes & 102 \\
\bottomrule
\end{tabular}
\caption{Overall search and reasoning statistics on LoCoMo.}
\label{tab:overall_stats}
\end{table}
\begin{table}[h!]
\centering
\small
\begin{tabular}{lrr}
\toprule
\textbf{Path Length} & \textbf{Count} & \textbf{Percentage} \\
\midrule
1  & 172 & 11.2\% \\
2  & 410 & 26.6\% \\
3  & 336 & 21.8\% \\
4  & 217 & 14.1\% \\
5  & 155 & 10.1\% \\
6  & 169 & 11.0\% \\
7  &  21 &  1.4\% \\
8  &  32 &  2.1\% \\
9  &  18 &  1.2\% \\
10 &   6 &  0.4\% \\
11 &   4 &  0.3\% \\
\bottomrule
\end{tabular}
\caption{Distribution of exploration path lengths in memory search.}
\label{tab:path_length}
\end{table}

\subsection{Per-Item Aggregated Statistics}
Table~\ref{tab:item_stats} reports statistics aggregated by item groups in LoCoMo. Across different items, the average runtime and exploration depth remain relatively stable, suggesting that the proposed search mechanism adapts robustly to different dialogue structures and content distributions. Variations in refinement rate and retained evidence reflect differences in reasoning complexity across items, rather than instability in the search process.

\begin{table*}[h!]
\centering
\small
\begin{tabular}{lrrrrr}
\toprule
\textbf{Item} & \textbf{\#Q} & \textbf{Avg.\ Time (s)} & \textbf{Avg.\ Steps} & \textbf{Refine \%} & \textbf{Avg.\ Kept} \\
\midrule
locomo\_item1  & 152 & 21.73 & 7.9 & 82.9 & 3.3 \\
locomo\_item2  &  81 & 21.21 & 7.6 & 80.2 & 3.1 \\
locomo\_item3  & 152 & 20.02 & 6.9 & 79.6 & 2.8 \\
locomo\_item4  & 199 & 22.28 & 8.5 & 71.9 & 3.7 \\
locomo\_item5  & 178 & 19.57 & 6.9 & 77.5 & 2.3 \\
locomo\_item6  & 123 & 21.25 & 7.8 & 77.2 & 3.4 \\
locomo\_item7  & 150 & 17.89 & 5.7 & 80.7 & 2.0 \\
locomo\_item8  & 191 & 20.05 & 7.1 & 70.7 & 3.1 \\
locomo\_item9  & 156 & 21.96 & 8.0 & 76.9 & 3.6 \\
locomo\_item10 & 158 & 22.80 & 8.8 & 70.9 & 4.0 \\
\bottomrule
\end{tabular}
\caption{Aggregated search statistics per item group.}
\label{tab:item_stats}
\end{table*}

\subsection{Statistics by Question Category}

Table~\ref{tab:category_stats} summarizes the search and reasoning behavior of \ours{} across different question categories. Reasoning-intensive questions, particularly multi-hop, require longer search trajectories, as reflected by higher average steps and longer processing time. Temporal questions, while involving fewer steps on average, exhibit the highest refinement rate, indicating frequent use of query refinement to resolve temporal dependencies. Single-hop questions are generally easier, requiring fewer steps and refinements while maintaining a high subgoal satisfaction rate. Overall, these patterns align well with the inherent complexity of each category and suggest that \ours{} adapts its search behavior according to task demands.

\begin{table*}[h!]
\centering
\small
\begin{tabular}{lrrrrr}
\toprule
\textbf{Category} & \textbf{\#Q} & \textbf{Avg.\ Time (s)} & \textbf{Avg.\ Steps} & \textbf{Subgoal Sat.\ \%} & \textbf{Refine \%} \\
\midrule
Multi-hop   & 282 & 24.61 & 10.1 & 71.5 & 78.7 \\
Temporal   & 321 & 18.64 & 5.9  & 61.3 & 83.8 \\
Open-domain   & 96 & 24.49 & 9.2  & 57.9 & 85.4 \\
Single-hop   & 841 & 20.05 & 7.1  & 71.0 & 71.7 \\
\bottomrule
\end{tabular}
\caption{Search and reasoning statistics by question category.}
\label{tab:category_stats}
\end{table*}

\subsection{Path Length Distribution}
Table~\ref{tab:path_length} shows the distribution of exploration path lengths. Most paths are short, with the majority falling between two and four steps, indicating that useful evidence is typically reached through localized reasoning over event relations. Longer paths are rare and correspond to more complex queries requiring extended exploration, demonstrating that deep traversal is selectively invoked rather than pervasive.

\section{Case Study: Multi-hop Reasoning over the Event Graph}
\label{sec:case_study}

We present a representative multi-hop question from the LoCoMo benchmark to qualitatively illustrate how \ours{} performs logic-aware memory search and reasoning over the Event Graph. This example highlights how evidence is incrementally constructed through structured traversal rather than flat retrieval. The case query is:
\begin{quote}
\emph{``What kinds of artworks did the speaker mention creating after moving to the new city?''}
\end{quote}

Answering this question requires linking events about relocation with later creative activities that are mentioned in separate dialogue segments.

\paragraph{Planner: Subgoal Decomposition}
Given the query, the Planner decomposes it into three subgoals:
\begin{itemize}
    \item $h_1$: Identify the event describing the speaker's move to a new city.
    \item $h_2$: Find events mentioning artistic or creative activities after the move.
    \item $h_3$: Extract the specific types of artworks mentioned.
\end{itemize}

The Planner initializes the subgoal satisfaction vector as $\mathbf{s} = [0, 0, 0]$, which is updated as evidence is collected.

\paragraph{Localization: Selecting Starting Events}
Using embedding similarity, the system retrieves the top-$5$ candidate events and selects starting nodes from $5$ distinct topic clusters. Example starting events include:
\begin{quote}
\emph{``Moved to Chicago last summer for a new job.''} \\
\emph{``I have been spending weekends exploring art museums.''}
\end{quote}

A total of $3$ starting nodes are inserted into the global exploration queue.

\paragraph{Explorer: Multi-path Navigation and Evidence Collection}
Three Explorer agents traverse the Event Graph in parallel. At each visited event, the Explorer conditions on the query, current subgoals, retained evidence, and local graph relations. The retained evidence set is updated accordingly. Across all paths, the agent explores $10$ candidate nodes, retains $7$ as evidence, and reaches an average path length of $2.84$ steps.

\paragraph{Query Refinement}
After the first exploration round, the Planner observes that $h_3$ is only partially supported. It triggers a single refinement step to focus on missing details:
\begin{quote}
\emph{``What specific forms of art did the speaker create after moving to the new city?''}
\end{quote}

This refined query guides a second round of targeted exploration, a mechanism triggered in $76.4\%$ of LoCoMo questions overall.

\paragraph{Evidence Aggregation}
After refinement, the retained evidence set consists of the following key events:
\begin{quote}
\emph{``Moved to Chicago last summer for a new job.''} \\
\emph{``I started painting landscapes in my apartment.''} \\
\emph{``I also experimented with stained glass designs.''}
\end{quote}

These events jointly satisfy all subgoals, yielding $\mathbf{s} = [1, 1, 1]$.

\paragraph{Answer Generation}
The Answerer generates the final response conditioned only on the distilled evidence:
\begin{quote}
\emph{``The speaker created paintings and stained glass artworks after moving.''}
\end{quote}

\paragraph{Discussion.}
This case illustrates how \ours{} constructs answers through guided traversal over logically connected events. Rather than retrieving a single text chunk, the agent incrementally accumulates evidence across multiple paths, refines its search when gaps are detected, and reasons over event dependencies. This process mirrors human multi-step recall and demonstrates the advantage of event-centric memory for complex multi-hop reasoning.

\section{Use of AI Assistants}

We use ChatGPT to improve the presentations of this paper.\footnote{\url{https://chatgpt.com/}}

\clearpage
\onecolumn
\section{Prompt Templates}
\label{append:prompt}
\subsection{Memory Construction}
\label{append:construction-prompt}
\begin{tcolorbox}[
    enhanced,
    breakable,        
    title=Event Extraction Prompt,
    colback=gray!2!white,
    colframe=gray!75!black,
    fonttitle=\bfseries,
    width=\columnwidth,   
    left=3pt, right=3pt,  
    top=5pt, bottom=5pt,
    boxsep=1pt,
    fontupper=\small      
]
    \noindent You are an expert information extraction system. Given a multi-turn dialog, extract meaningful events and output ONE strict JSON object.

    \medskip
    \noindent\textbf{Goals:}
    \par \textbullet\ Extract logically coherent events (E1, E2, ...) in chronological order. Each event represents a complete logical unit.
    \par \textbullet\ \textbf{AGGRESSIVELY COMBINE} related micro-events into comprehensive summaries to avoid fragmentation. Merge events that:
    \par \hspace*{1em} -- Involve same participants discussing the same topic.
    \par \hspace*{1em} -- Form a logical sequence (decision + action + completion).
    \par \hspace*{1em} -- Are temporally close and thematically related (within 3-5 utterances).
    \par \hspace*{1em} -- Represent different aspects of the same situation/problem.
    \par \hspace*{1em} -- Include follow-up questions, clarifications, or elaborations.

    \smallskip
    \par \textbullet\ \textbf{PRESERVE ALL} important details within each merged event summary. Include:
    \par \hspace*{1em} -- Complete context and all key outcomes, results, and conclusions.
    \par \hspace*{1em} -- Specific facts, numbers, dates, locations, and concrete details.
    \par \hspace*{1em} -- Emotional states, reactions, and interpersonal dynamics.
    \par \hspace*{1em} -- Technical details, requirements, and specifications.
    \par \hspace*{1em} -- Any conditions, constraints, or limitations discussed.
    \par \hspace*{1em} -- \textbf{IMPORTANT}: Include visual content descriptions for shared images.

    \smallskip
    \par \textbullet\ \textbf{Constraints}: List people involved as an array \texttt{people} (max 3). Do not output other entity types or attributes.
    \par \textbullet\ \textbf{Event Count}: Extract 6-10 comprehensive events. Prioritize fewer, more detailed events over many fragmented ones.

    \medskip
    \hrule
    \medskip

    \noindent\textbf{RESPONSE FORMAT:}
    \par\noindent Output JSON only, no additional commentary.
\end{tcolorbox}

\begin{tcolorbox}[
    enhanced,
    breakable,            
    title=Event Relation Extraction Prompt,
    colback=gray!2!white,
    colframe=gray!75!black,
    fonttitle=\bfseries,
    width=\columnwidth,   
    left=3pt, right=3pt,  
    top=5pt, bottom=5pt,
    boxsep=1pt,
    fontupper=\small      
]
    \noindent You are an expert information extraction system. Given a list of extracted events from a dialog, identify meaningful pairwise relations between them and output ONE strict JSON object.

    \medskip
    \noindent\textbf{Goals:}
    \par \textbullet\ Consider \textbf{ALL} unordered pairs of events within the same session (not only adjacent events).
    \par \textbullet\ Extract pairwise event relations with a SHORT, free-form label in \texttt{type} that best characterizes the link.
    \par \textbullet\ Relation types can include: \textit{causal, motivation, enablement, follow\_up, temporal\_before, temporal\_after, contrast, part\_of, parallel, elaboration}. These are examples, not a closed set.
    \par \textbullet\ Add relations only when meaningful. Prefer specific semantic links over trivial temporal ordering.
    \par \textbullet\ It is acceptable to have no temporal edges if they add no insight.

    \medskip
    \noindent\textbf{CRITICAL GUIDELINES:}
    \par \textbullet\ \textbf{IMPORTANT}: For temporal relations (\textit{follow\_up, temporal\_before, temporal\_after}), base them on the \textbf{ACTUAL TIME} when events occurred in the real world, NOT on when they are described in the dialog. Focus on the chronological sequence of reality.
    \par \textbullet\ For each relation, cite minimal \texttt{evidence} utterance ids that support the linkage between the two events.

    \medskip
    \hrule
    \medskip

    \noindent\textbf{RESPONSE FORMAT:}
    \par\noindent Output JSON only, no additional commentary.
\end{tcolorbox}

\begin{tcolorbox}[
    enhanced,
    breakable,            
    title=Event Coreference \& Overlap Prompt,
    colback=gray!2!white,
    colframe=gray!75!black,
    fonttitle=\bfseries,
    width=\columnwidth,   
    left=3pt, right=3pt,  
    top=5pt, bottom=5pt,
    boxsep=1pt,
    fontupper=\small      
]
    \noindent You are an expert at analyzing events and determining if they refer to the same real-world occurrence or have significant overlap.

    \medskip
    \noindent Given two event descriptions extracted from different dialog sessions, determine:
    \par \textbullet\ 1. Whether they describe the \textbf{SAME} event (same occurrence at the same time).
    \par \textbullet\ 2. Whether they have \textbf{SIGNIFICANT OVERLAP} (mention or relate to the same real-world situation/topic).

    \medskip
    \noindent\textbf{Consider these factors:}
    \par \textbullet\ Do they involve the same people/participants?
    \par \textbullet\ Do they describe the same actions, situations, or topics?
    \par \textbullet\ Do they have compatible time references?
    \par \textbullet\ Would merging their information create a more complete picture of ONE event?

    \medskip
    \hrule
    \medskip

    \noindent\textbf{Output a JSON object with these exact keys:}
    \par\smallskip
    \noindent \texttt{\{}
    \par\indent \texttt{"same\_event": boolean,} \hfill // true if they are the same event
    \par\indent \texttt{"has\_overlap": boolean,} \hfill // true if they refer to the same situation
    \par\indent \texttt{"relation\_type": string | null,} \hfill // suggest relation type if overlap
    \par\indent \texttt{"reasoning": string} \hfill // brief explanation
    \par\noindent \texttt{\}}
\end{tcolorbox}

\subsection{Memory Search}
\label{append:search-prompt}
\begin{tcolorbox}[
    enhanced,
    breakable,            
    title=Action Decision Prompt,
    colback=gray!2!white,
    colframe=gray!75!black,
    fonttitle=\bfseries,
    width=\columnwidth,   
    left=3pt, right=3pt,  
    top=5pt, bottom=5pt,
    boxsep=1pt,
    fontupper=\small      
]
    \noindent You are an expert information evaluator. Your task is to decide which action to take for the current node based on how relevant and sufficient it is for answering the given question. You have \textbf{THREE} possible actions:

    \medskip
    \noindent\textbf{1. SKIP}: The current node is NOT helpful for answering the question or satisfying any sub-goals.
    \par \textbullet\ Use SKIP when the current node contains completely irrelevant information.
    \par \textbullet\ The current node will be DISCARDED, not used in final answer.
    \par \textbullet\ You should specify which neighbor node(s) to explore next, OR specify NONE if ALL neighbors are irrelevant.
    \par \textbullet\ \textbf{Multi-node selection rules}: Maximum 3 nodes, only select HIGHLY relevant ones.

    \medskip
    \noindent\textbf{2. EXPAND}: The current node IS helpful and helps satisfy some sub-goals, but NOT all sub-goals are satisfied yet.
    \par \textbullet\ Use EXPAND when the current node contains useful information for one or more sub-goals.
    \par \textbullet\ The current node will be KEPT and used in the final answer.
    \par \textbullet\ Specify neighbor node(s) to explore next to satisfy remaining sub-goals, OR specify NONE if no neighbors are relevant.
    \par \textbullet\ \textbf{CRITICAL}: You MUST indicate which sub-goals are now satisfied by this node + previously kept information. Only mark a sub-goal as satisfied if you have DIRECT evidence.

    \medskip
    \noindent\textbf{3. ANSWER}: Use ONLY when ALL sub-goals are SATISFIED (or nearly all).
    \par \textbullet\ Use ANSWER when the previously kept information + current node together satisfy ALL sub-goals.
    \par \textbullet\ The current node will be KEPT and exploration will STOP.
    \par \textbullet\ \textbf{CRITICAL}: You MUST list ALL satisfied sub-goals to confirm completeness.
    \par \textbullet\ Be conservative: If ANY sub-goal remains unsatisfied, use EXPAND instead.

    \medskip
    \hrule
    \medskip

    \noindent\textbf{CRITICAL GUIDELINES:}
    \par \textbf{-- Check sub-goals systematically}: For each action, explicitly evaluate which sub-goals are satisfied.
    \par \textbf{-- ANSWER only when complete}: Use ANSWER only when ALL (or all critical) sub-goals are satisfied.
    \par \textbf{-- Navigate strategically}: Choose next nodes that are likely to help satisfy remaining unsatisfied sub-goals.
    \par \textbf{-- Be explicit about progress}: Always indicate which sub-goals your current decision addresses.

    \medskip
    \noindent\textbf{RESPONSE FORMAT (follow strictly):}
    \par\noindent ACTION: [SKIP/EXPAND/ANSWER]
    \par\noindent NEXT\_NODES: [NODE\_ID1, NODE\_ID2, ...] (or NONE)
    \par\noindent SATISFIED\_SUBGOALS: [1, 3, 4] (REQUIRED for EXPAND/ANSWER; [] for SKIP)
    \par\noindent REASONING: [Brief explanation: (1) info provided, (2) sub-goals satisfied, (3) sub-goals remaining, (4) why chosen next nodes target remaining sub-goals]

    \medskip
    \noindent\textbf{IMPORTANT}: (1) For SKIP, SATISFIED\_SUBGOALS must be []; (2) For EXPAND/ANSWER: provide list even if empty; (3) Only include sub-goals with DIRECT evidence; (4) Do NOT speculate.

    \medskip
    \hrule
    \medskip

    \noindent QUESTION: \{question\}
    \par \{subgoals\_text\}

    \medskip
    \noindent PREVIOUSLY KEPT INFORMATION:
    \par \{kept\_nodes\_info if kept\_nodes\_info else "(No information kept yet)"\}

    \medskip
    \noindent CURRENT NODE INFORMATION:
    \par \{current\_info\}

    \medskip
    \noindent NEIGHBOR NODES (available for exploration):
    \par \{neighbor\_info\}

    \medskip
    \noindent\textbf{Now, make your decision:}
\end{tcolorbox}

\begin{tcolorbox}[
    enhanced,
    breakable,           
    title=Response Generation Prompt,
    colback=gray!2!white,
    colframe=gray!75!black,
    fonttitle=\bfseries,
    width=\columnwidth,   
    left=4pt, right=4pt,  
    top=5pt, bottom=5pt,
    boxsep=1pt,
    fontupper=\small      
]
    \noindent Your task is to answer the QUESTION based on the provided CONTEXT.

    \medskip
    \noindent\textbf{Requirements:}
    \par \textbullet\ \textbf{Be concise and direct}: Provide ONLY the answer in the form of \textbf{a short phrase}, not a sentence. No explanations or additional commentary.
    \par \textbullet\ \textbf{Original wording}: If the context contains direct statements that answer the question, use the original wording from the context.
    \par \textbullet\ \textbf{Inference}: If the context doesn't have direct statements, you may summarize and infer the answer from the relevant information.
    \par \textbullet\ \textbf{Time Reference Calculation}: If there is a question about time references (like "last year", "two months ago", etc.), calculate the actual date based on the memory timestamp. 
    \par \textit{Example}: If a memory from 4 May 2022 mentions "went to India last year," then the trip occurred in 2021.
    \par \textbullet\ \textbf{Specific Dates}: Always convert relative time references to specific dates, months, or years. For example, convert "last year" to "2022" or "two months ago" to "March, 2023" based on the memory timestamp.
    \par \textbullet\ \textbf{Reasonable Justification}: If you are uncertain or lack sufficient information, do not state that the information is insufficient. Instead, provide a reasonable and well-justified answer based on general knowledge.
    \par \textbullet\ \textbf{Keep it brief}: Keep your answer brief and to the point.

    \medskip
    \hrule
    \medskip

    \noindent\textbf{CONTEXT:}
    \par \{context\}

    \medskip
    \noindent\textbf{QUESTION:}
    \par \{question\}

    \medskip
    \noindent\textbf{ANSWER:}
\end{tcolorbox}

\begin{tcolorbox}[
    enhanced,
    breakable,            
    title=Refinement Query Prompt,
    colback=gray!2!white,
    colframe=gray!75!black,
    fonttitle=\bfseries,
    width=\columnwidth,   
    left=4pt, right=4pt,  
    top=5pt, bottom=5pt,
    boxsep=1pt,
    fontupper=\small      
]
    \noindent You are an assistant whose role is to generate a refined search query to find missing information.

    \medskip
    \noindent\textbf{ORIGINAL QUESTION:}
    \par \{original\_question\}

    \medskip
    \noindent\textbf{SUB-GOALS STATUS:}
    \par \noindent Satisfied sub-goals:
    \par \{satisfied\_text\}
    \par \noindent Unsatisfied sub-goals:
    \par \{unsatisfied\_text\}

    \medskip
    \noindent\textbf{INFORMATION COLLECTED SO FAR:}
    \par \{context\_so\_far\}

    \medskip
    \hrule
    \medskip

    \noindent\textbf{TASK:}
    \par Generate a NEW search query that specifically targets the UNSATISFIED sub-goals.

    \medskip
    \noindent\textbf{Your new query should:}
    \par \textbullet\ 1. Focus on the specific unsatisfied sub-goals.
    \par \textbullet\ 2. Be clear and specific.
    \par \textbullet\ 3. Use different keywords or phrases than the original question.
    \par \textbullet\ 4. Target information that would help satisfy the remaining sub-goals.
    \par \textbullet\ 5. NOT repeat the original question.

    \medskip
    \hrule
    \medskip

    \noindent\textbf{RESPONSE FORMAT:}
    \par \noindent New Query: [Your refined search query - single clear question or search phrase targeting unsatisfied sub-goals]
    \par \noindent Target Sub-goals: [List which sub-goal numbers this query aims to satisfy]

    \medskip
    \noindent\textbf{Generate your response:}
\end{tcolorbox}

\begin{tcolorbox}[
    enhanced,
    breakable,            
    title=Memory Node Selection Prompt,
    colback=gray!2!white,
    colframe=gray!75!black,
    fonttitle=\bfseries,
    width=\columnwidth,   
    left=3pt, right=3pt,  
    top=5pt, bottom=5pt,
    boxsep=1pt,
    fontupper=\small      
]
    \noindent You are selecting the most promising memory nodes to explore for answering a question.

    \medskip
    \noindent\textbf{QUESTION:} \{question\}
    \par \{subgoals\_text\}

    \medskip
    \noindent\textbf{CANDIDATE NODES (retrieved by semantic similarity):}
    \par \{nodes\_text\}

    \medskip
    \hrule
    \medskip

    \noindent\textbf{INSTRUCTIONS:}
    \par Select the nodes that are HIGHLY LIKELY to contain information relevant to one or more sub-goals.
    \par \textbullet\ \textbf{Be selective}: Only choose nodes whose summaries clearly indicate relevance to specific sub-goals.
    \par \textbullet\ \textbf{Maximum 5 nodes}: Select at most 5 nodes to explore.
    \par \textbullet\ \textbf{Diversity}: Try to select nodes that address different sub-goals if possible.
    \par \textbullet\ \textbf{Quality over quantity}: It's better to select 2 highly relevant nodes than 5 marginally relevant ones.
    \par \textbullet\ If a node's summary is vague or doesn't clearly relate to any sub-goal, DON'T select it.
    \par \textbullet\ Consider both the summary content and the similarity score.

    \medskip
    \hrule
    \medskip

    \noindent\textbf{RESPONSE FORMAT:}
    \par\noindent Selected Nodes: [NODE\_ID1, NODE\_ID2, ...]
    \par\noindent Reasoning: [Brief explanation of why each selected node is likely relevant to specific sub-goals]

    \medskip
    \noindent\textbf{Now make your selection:}
\end{tcolorbox}

\begin{tcolorbox}[
    enhanced,
    breakable,            
    title=Cluster-based Node Selection Prompt,
    colback=gray!2!white,
    colframe=gray!75!black,
    fonttitle=\bfseries,
    width=\columnwidth,   
    left=3pt, right=3pt,  
    top=5pt, bottom=5pt,
    boxsep=1pt,
    fontupper=\small      
]
    \noindent You are selecting the most relevant memory node(s) to answer a question.

    \medskip
    \noindent\textbf{QUESTION:} \{question\}

    \medskip
    \noindent\textbf{AVAILABLE NODES:}
    \par \{nodes\_text\}

    \medskip
    \hrule
    \medskip

    \noindent\textbf{INSTRUCTIONS:}
    \par Select the node(s) that are HIGHLY relevant to answering the question.
    \par \textbullet\ \textbf{Be selective}: Only choose nodes that are HIGHLY relevant to the question.
    \par \textbullet\ \textbf{Maximum 3 nodes}: Select at most 3 nodes per cluster.
    \par \textbullet\ If ONLY ONE node is clearly the most relevant, select just that one.
    \par \textbullet\ Select multiple nodes (2-3) ONLY when they are ALL highly relevant AND provide complementary information:
    \par \hspace*{1em} * Information is distributed across multiple memories about the SAME topic.
    \par \hspace*{1em} * The question has multiple specific aspects that DIFFERENT nodes address.
    \par \hspace*{1em} * Multiple nodes provide different pieces of the SAME answer.
    \par \textbullet\ Consider the summary content, people involved, and time information.
    \par \textbullet\ \textbf{Do NOT select nodes that are only tangentially related or vaguely relevant.}

    \medskip
    \hrule
    \medskip

    \noindent\textbf{RESPONSE FORMAT:}
    \par\noindent Selected Nodes: [NODE\_ID1, NODE\_ID2, ...] 
    \par\noindent Reason: [Brief explanation of why these specific nodes are HIGHLY relevant]
\end{tcolorbox}

\begin{tcolorbox}[
    enhanced,
    breakable,            
    title=Strategic Planning Prompt,
    colback=gray!2!white,
    colframe=gray!75!black,
    fonttitle=\bfseries,
    width=\columnwidth,  
    left=3pt, right=3pt,  
    top=5pt, bottom=5pt,
    boxsep=1pt,
    fontupper=\small      
]
    \noindent You are a strategic planning assistant. Your task is to analyze a question and break it down into 2-5 specific sub-goals that need to be satisfied to fully answer the question.

    \medskip
    \noindent\textbf{QUESTION:} \{question\}

    \medskip
    \hrule
    \medskip

    \noindent\textbf{INSTRUCTIONS:}
    \par \textbullet\ 1. Analyze what information components are needed to fully answer this question.
    \par \textbullet\ 2. Break down the question into 2-5 specific, concrete sub-goals.
    \par \textbullet\ 3. Each sub-goal should represent a distinct piece of information needed.
    \par \textbullet\ 4. Sub-goals should be:
    \par \hspace*{1em} -- Specific and clear (not vague)
    \par \hspace*{1em} -- Independently verifiable (can determine if it's satisfied)
    \par \hspace*{1em} -- Collectively sufficient (together they fully answer the question)
    \par \hspace*{1em} -- Atomic (each sub-goal addresses ONE aspect)

    \medskip
    \hrule
    \medskip

    \noindent\textbf{RESPONSE FORMAT (follow strictly):}
    \par\noindent Sub-goal 1: [First specific information need]
    \par\noindent Sub-goal 2: [Second specific information need]
    \par\noindent Sub-goal 3: [Third specific information need]
    \par\noindent ...

    \medskip
    \noindent\textbf{Now analyze the question and generate sub-goals:}
\end{tcolorbox}

\end{document}